\documentclass{article}
\usepackage{ijcai25}
\usepackage{listings}
\usepackage{booktabs}
\usepackage{threeparttable}
\usepackage{times}
\usepackage{soul}
\usepackage{url}
\usepackage[hidelinks]{hyperref}
\usepackage[utf8]{inputenc}
\usepackage[small]{caption}
\usepackage{graphicx}
\usepackage{amsmath}
\usepackage{amsthm}
\usepackage{booktabs}
\usepackage[switch]{lineno}

\usepackage{amsmath}
\usepackage{amssymb}
\usepackage{amsfonts}      
\usepackage[ruled,vlined]{algorithm2e}
\usepackage{booktabs}
\usepackage{multirow}
\usepackage{float}
\SetArgSty{textup} 
\usepackage{fancyvrb}
\usepackage[most]{tcolorbox}
\usepackage{spverbatim}

\title{AriGraph: Learning Knowledge Graph World Models with Episodic Memory for LLM Agents}

\author{
Petr Anokhin$^1$
\and
Nikita Semenov$^2$\and
Artyom Sorokin$^1$\and
Dmitry Evseev$^2$\and
Andrey Kravchenko$^4$\and
Mikhail Burtsev$^3$\And
Evgeny Burnaev$^{2,1}$\\
\affiliations
$^1$AIRI, Moscow, Russia\\
$^2$Skoltech, Moscow, Russia\\
$^3$London Institute for Mathematical Sciences, London, UK\\
$^4$University of Oxford, UK\\
\emails
anokhin@airi.net
}

\begin{document}

\maketitle

\begin{abstract}
    Advancements in the capabilities of Large Language Models (LLMs) have created a promising foundation for developing autonomous agents. With the right tools, these agents could learn to solve tasks in new environments by accumulating and updating their knowledge. Current LLM-based agents process past experiences using a full history of observations, summarization, retrieval augmentation. However, these unstructured memory representations do not facilitate the reasoning and planning essential for complex decision-making. In our study, we introduce AriGraph, a novel method wherein the agent constructs and updates a memory graph that integrates semantic and episodic memories while exploring the environment. We demonstrate that our Ariadne LLM agent, consisting of the proposed memory architecture augmented with planning and decision-making, effectively handles complex tasks within interactive text game environments difficult even for human players. Results show that our approach markedly outperforms other established memory methods and strong RL baselines in a range of problems of varying complexity. Additionally, AriGraph demonstrates competitive performance compared to dedicated knowledge graph-based methods in static multi-hop question-answering. 
\end{abstract}

\section{Introduction}

\begin{figure*}[t]
  \centering
  \includegraphics[width=0.9\textwidth]{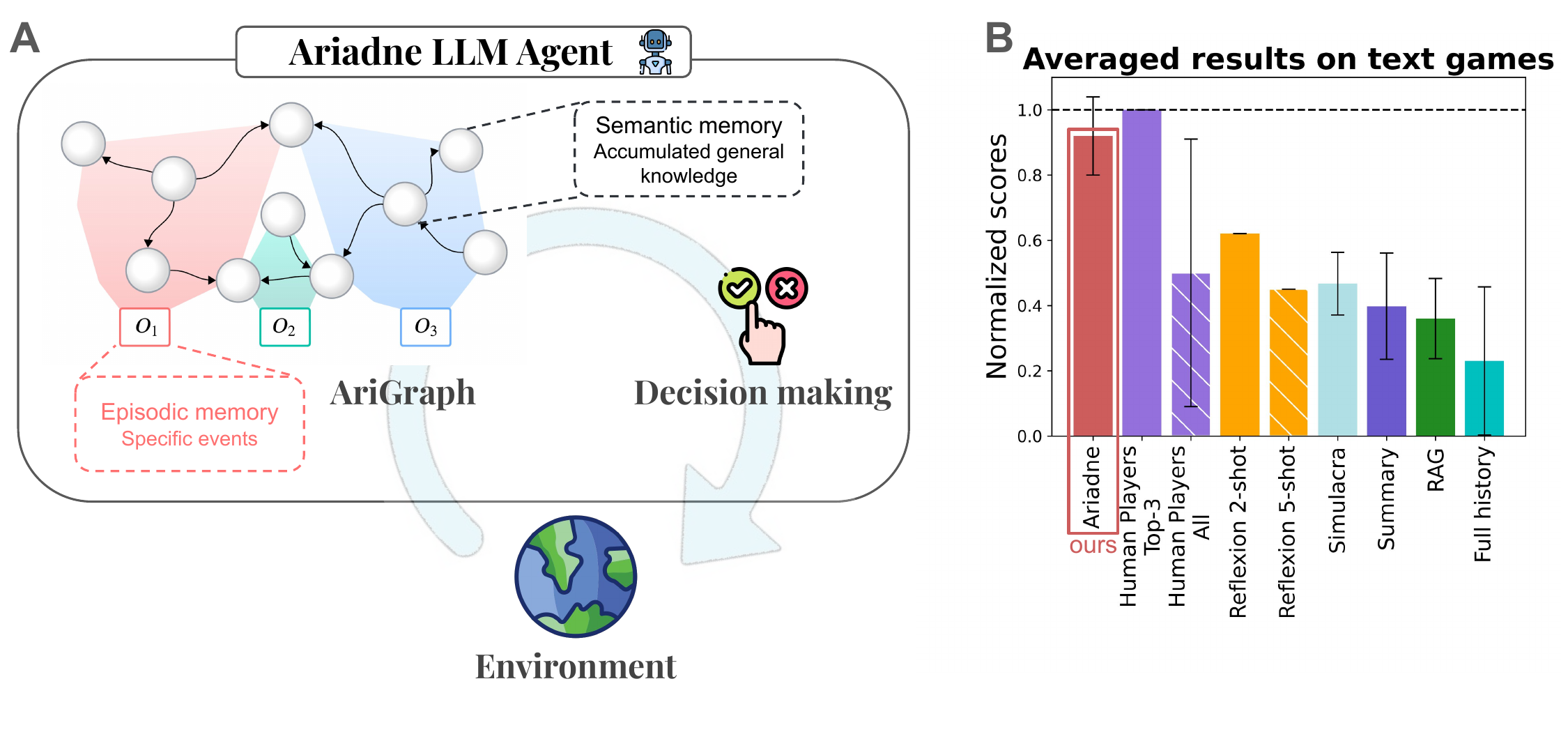}
  \caption{ (A) The architecture of our Ariadne agent, equipped with AriGraph memory. AriGraph integrates both semantic knowledge graph and past experiences. Memory in the form of a semantic knowledge graph extended with episodic vertices and edges significantly enhances the performance of LLM-agent in text-based games. 
(B) The average performance of our agent on text games, compared to various baselines including human players and other LLM memory implementations. The LLM-agents differ only in the memory module, while the decision-making component remains identical across all versions. The results for the agents are displayed for the top three out of five runs. For human players, the results are presented as both the top three and the average across all participants.}
  \label{fig:agent_arch}
\end{figure*}

Impressive language generation capabilities of large language models (LLMs) has sparked substantial interest in their application as core components for creating autonomous agents capable of interacting with dynamic environments and executing complex tasks. Over the past year, the research community has explored general architectures and core modules for such LLM agents \cite{Wang_2024,sumers2024cognitive,cheng2024exploring}.  A crucial property of a general cognitive agent is its ability to accumulate and use knowledge. A long-term memory allows an agent to store and recall past experiences and knowledge, enabling it to learn from previous encounters and make informed decisions. However, the question of the best way to equip an agent with these capabilities remains open. Despite the constraints inherent in transformer architectures, contemporary methods enable LLMs to manage contexts encompassing millions of tokens~\cite{ding2024longrope}. However, this approach proves inefficient for agents required to maintain continuous interaction with their environment. Such agents must hold an entire historical context in memory to perform actions, which is not only costly but also limited in handling complex logic hidden in vast amounts of information. Research into alternative frameworks like Recurrent Memory Transformer \cite{bulatov2022recurrent,bulatov2024scaling}  and MAMBA \cite{gu2023mamba} seeks to provide long-term memory solutions, though these models are still in their infancy.

Currently, the most popular solution for incorporating memory to LLM agents is the Retrieval-Augmented Generation (RAG) approach. RAG in a form of vector retrieval leverages an external database to enhance the model's prompt with relevant information. This technique is commonly used in memory architectures for LLM agents, often to recall specific observations or learned skills. However, it suffers from unstructured nature, greatly reducing the ability to retrieve related information, which may be scattered throughout the agent's memory. These limitations can be overcome by using knowledge graphs as database. This approach has also experienced a resurgence with the advent of LLMs \cite{Pan_2024}.  However, for a robust memory architecture, integrating both structured and unstructured data is essential. In cognitive science, this integration parallels the concepts of semantic and episodic memories. Semantic memory encompasses factual knowledge about the world, whereas episodic memory pertains to personal experiences, which often contain richer and more detailed information. Though traditionally considered separate due to their distinct neurological representations, recent studies suggest these memory types are interconnected \cite{WongGonzalez2018}. Semantic knowledge is built upon the foundation of episodic memory and subsequently provides a structured base for associative memory. This allows for the integration of various memory aspects, including episodic memories themselves.

In our research, we have developed a memory architecture called Ariadne's Graph (AriGraph), that integrates semantic and episodic memories within a memory graph framework. A knowledge graph represents a network of interconnected semantic knowledge, while episodic memories are depicted as episodic edges that can connect multiple relations within the graph. As an agent interacts with environment, it learns joint semantic and episodic world model by updating and extending knowledge graph based memory. 
This architecture not only serves as a foundational memory framework but also aids in environmental modeling, improving spatial orientation and exploration capabilities. For the general framework of our LLM agent called Ariadne, we employed pipeline of memory retrieval, planing and decision making. 
For evaluation of proposed methods we set up experiments to study two research questions.

\textbf{RQ1.} Can LLM based agents learn useful structured world model from scratch via interaction with an environment?

\textbf{RQ2.} Does structured knowledge representation improve retrieval of relevant facts from memory and enable effective exploration?

We evaluated our agent in complex interactive tasks in Textworld  and NetHack environments~\cite{cote18textworld,NethackLE_kuttler2020}.  
Experimental results demonstrate that our agent Ariadne can effectively learn through interactions with environment and significantly outperforms other memory approaches for LLMs such as full history, summarization, RAG, Simulacra~\cite{park2023generative} and Reflexion~\cite{shinn2023reflexion}. We also show that our method outperforms existing reinforcement learning (RL) baselines. We also evaluated our approach on the classical roguelike game NetHack, where our agent with local observations achieved scores comparable to an agent with ground-truth knowledge. Although AriGraph was originally designed for an agent interacting with the environment, it also demonstrates competitive performance on multi-hop question answering tasks.

\section{AriGraph World Model}
\label{sec:arigraph}

\begin{figure*}[t]
  \centering
  \includegraphics[width=0.49\textwidth]{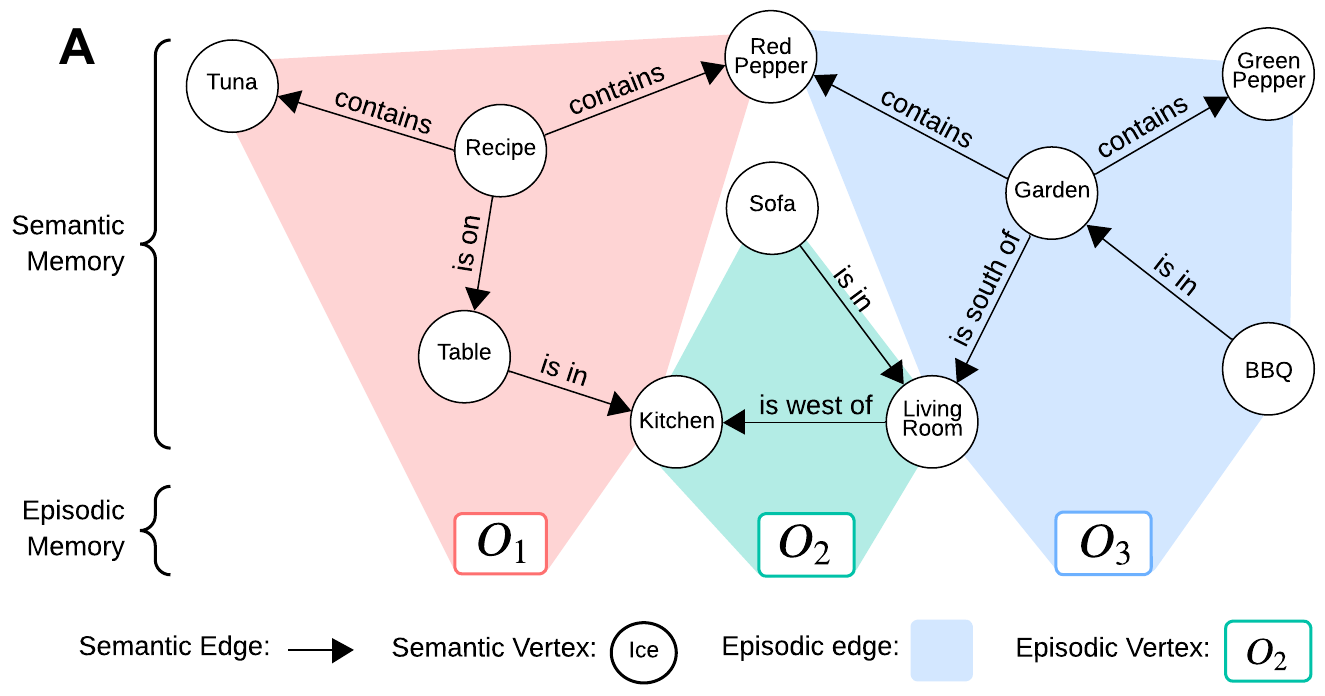}
  \includegraphics[width=0.5\textwidth]{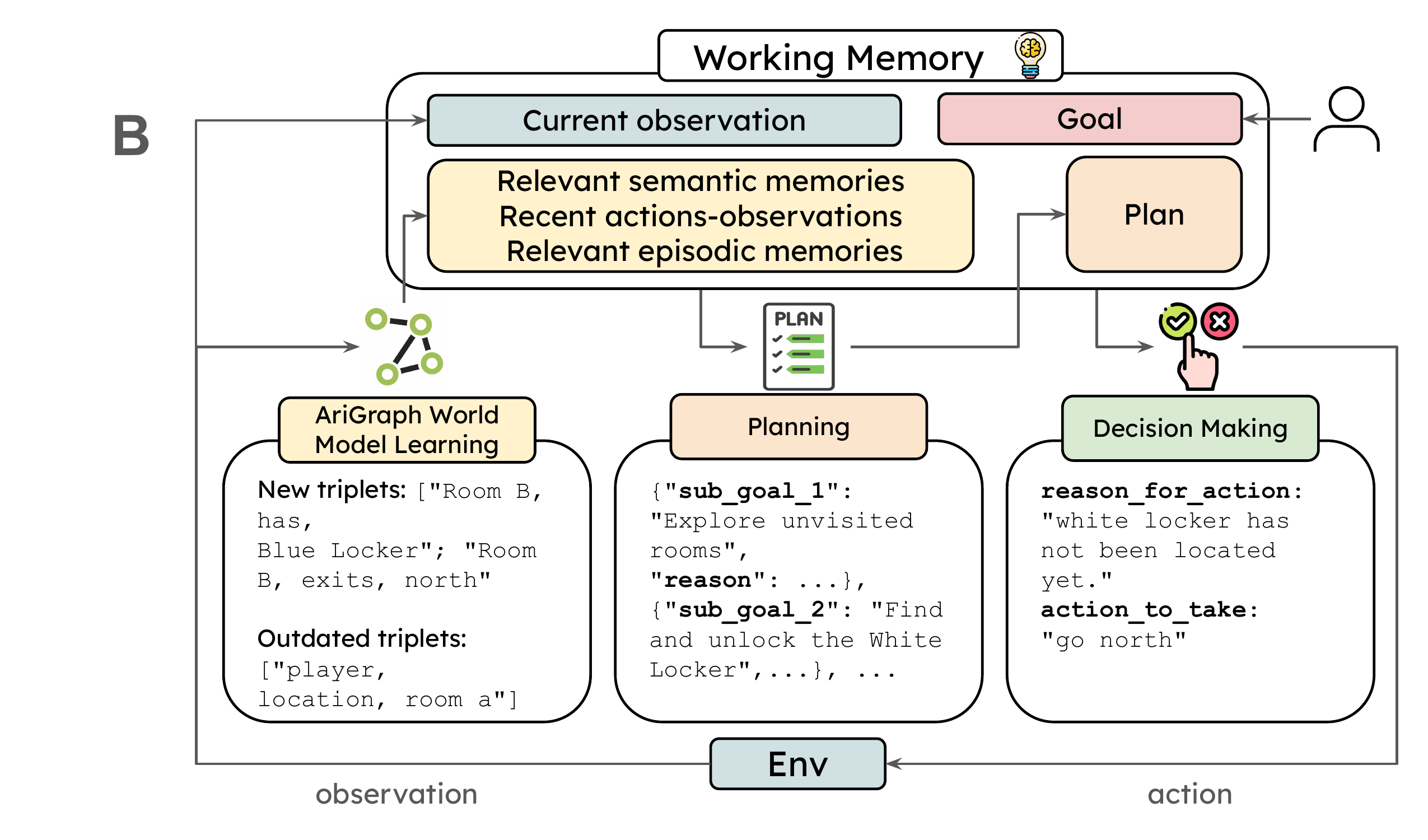}
  \caption{AriGraph world model and Ariadne cognitive architecture. (A) AriGraph learns episodic and semantic knowledge during interaction with unknown environment. At each time step $t$ new episodic vertex (containing full textual observation $o_t$) is added to the episodic memory. Then LLM model parses observation $o_t$ to extract relevant relationships in a form of triplets $(object_{1},\, relation,\, object_{2})$. These triplets are used to update semantic memory graph. The connection between episodic and semantic memory occurs through episodic edges that link each episodic vertex with all triplets extracted from respective observation. (B) Ariadne agent explores the environment and accomplishes tasks with AriGraph. User sets goal to the agent. Working memory is populated with recent history of observations and actions, relevant semantic and episodic knowledge retrieved from the AirGraph world model. Planing LLM module uses content of working memory to generate new or update existing plan. Results of planning are stored back in working memory.  Finally, a ReAct-based module reads memory content and selects one of possible actions to be executed in the environment. Every observation triggers learning that updates agent's world model.}
  \label{fig:graph_structure}
\end{figure*}

\textbf{Memory graph structure.} 
AriGraph world model \( G = (V_s, E_s, V_e, E_e) \) consists of  \textit{semantic} $(V_s, E_s)$ and \textit{episodic memory} $(V_e, E_e)$ vertices and edges (see Figure \ref{fig:graph_structure}). 
At each step $t$ agent receives observation $o_t$  and sends action $a_t$ back to the environment. 
The environment also returns rewards $r_t$ that are not visible to the LLM agent but are used to evaluate its performance.
The agent continuously learns world model $G$ by extracting semantic triplets \((object_1, relation, object_2)\) from textual observations $o_t$.

\begin{itemize}
    \item \( V_s \) is a set of semantic vertices. Semantic vertices correspond to objects extracted from triplets.
    \item \( E_s \) is a set of semantic edges. Semantic edge is a tuple $(v, rel, u)$, where $u$, $v$ are semantic vertices and $rel$ is a relationship between them. Semantic edges essentially represent triplets integrated in the \textit{semantic memory}.
    \item \( V_e \) is a set of episodic vertices. Each episodic vertex corresponds to an observation received from the environment at the respective step $v^t_e = o_t$.
    \item \( E_e \) is a set of episodic edges. Each episodic edge $e^t_e=(v^t_e, E^t_s)$ connects all semantic triplets $E^t_s$ extracted from $o_t$ with each other and corresponding episodic vertex $v^t_e$. In other words episodic edges represent temporal relationship ``happened at the same time''.\footnote{Strictly speaking, episodic edges cannot be called edges or even hyperedges, because they connect vertices with multiple graph edges, but for simplicity we call them edges or episodic edges.}    
\end{itemize}

\textbf{Constructing AriGraph.} 
Interaction with the environment can provide the agent with an information about the world to create new or update previously acquired knowledge. 
Given new observation $o_t$, LLM agent extracts new triplets as semantic vertices $V^{t}_s$ and edges $E^{t}_s$.
To find already existing knowledge about the objects mentioned in $o_t$ a set of all semantic edges $E^{rel}_s$ incident to vertices $V^{t}_s$ is filtered out. 
Then outdated edges in $E^{rel}_s$ are detected by comparing them with $E^t_s$ and removed from the graph.
After clearing outdated knowledge we expand semantic memory with $V^{t}_s$ and $E^t_s$.
Episodic memory is updated by simply adding new episodic vertex $v^t_e$ containing $o_t$ and new episodic edge that connect all edges in $E^t_s$ with $v^t_e$. 
Episodic nodes store agent's past history and episodic edges connects all knowledge received at the same step. See Appendix \ref{app:prompts} for prompts used to extract new triplets and detect outdated knowledge.

\textbf{Retrieval from AriGraph.}
For successful decision-making in a partially observable environment, the agent needs to be able to retrieve relevant knowledge. Retrieval from the AriGraph memory  consists of two procedures: (1) a semantic search  returns the most relevant triplets (semantic edges) and (2) an episodic search that, given extracted triplets, returns the most relevant episodic vertices $V_e$. The pseudo-code for the search is presented in the Algorithm \ref{algo:memory_graph_search}.

\begin{algorithm}[tb] 

    \SetAlgoLined
    \KwIn{set of queries $Q$, $V_s$, $E_s$, $V_e$, $E_e$, \\ \hspace{0.07\textwidth} number of episodic vertices $k$, semantic \\ \hspace{0.07\textwidth} search depth $d$ and width $w$}
    \KwResult{retrieved episodic vertices $V^Q_e$, retrieved semantic triplets $E^Q_s$}
    \
    $E^Q_s \leftarrow \emptyset$, \\ 
    \ForEach{$q$ 
    in
    $Q$}{
        $E'_s \leftarrow \texttt{SemanticSearch}(q, V_s, E_s, d, w)$ \\
        $E^Q_s \leftarrow E^Q_s \cup E'_s$ \\
    }
    $V^Q_e \leftarrow \texttt{EpisodicSearch}(E^Q_s, V_e, E_e, k)$  \\
    \Return{$E^Q_s$, $V^Q_e$}

\caption{Memory Graph Search}
\label{algo:memory_graph_search}
\end{algorithm}

Semantic search relies on semantic similarity and semantic graph structure to recall the most relevant triplets. 
Given a query, the retriever (pre-trained Contriever model \cite{izacard2022unsupervised}) selects the most relevant semantic triplets. Then, the set of vertices incident to the found triplets is used to recursively retrieve new edges from the graph. Depth and breadth of the search can be controlled by respective hyperparameters $d$ and $w$.  For details see Appendix \ref{app:memory_graph_search}.

Episodic search starts with the results of the semantic search as an input. Episodic edges link the input triplets with past episodic vertices representing observations. The number of input triplets associated with a particular episodic vertex is used to calculate  their relevance:
\begin{align}
    \label{eq:episodic_node_relevance}
    rel(v^i_e) = \frac{n_i}{max(N_i, 1)} \log(max(N_i, 1))\,,
\end{align}
where $n_i$ is a number of input triplets incident to episodic edge $e^i$, $N_i$ is a total number triplets~(semantic edges) incident to $e^i$ and $\log\ (max(N_i, 1))$ is a weighting factor to prevent high scores for low information observations. We apply a $log_2(N_i)$ scaling to give more weight to observations with more extracted triplets.  
Additionally, observations containing exactly one triplet are assigned zero weight, as they are unlikely to provide information beyond the triplet itself. 
$k$ most relevant episodic vertices (containing respective observations) are returned as a result of the episodic search.

\section{Ariadne cognitive architecture}
\label{sec:ariadne}

To test utility of AriGraph world modelling method we propose an agentic architecture called Ariadne. Ariadne agent interacts with an unknown environment to accomplish a goal set by a user. Throughout this process, at each time step, the agent learns a world model, plans and executes actions. Ariadne has \textit{long-term} memory stored as AriGraph and \textit{working} memory containing information for current planning and decision making. 

Given an observation the agent updates world model and retrieves semantic and episodic knowledge from  AriGraph to working memory. Working memory is also populated with a final goal description, current observation, history of recent observation and actions. At the planning stage, Ariadne agent uses content of working memory to create new or update existing plan as a series of task-relevant sub-goals, each accompanied by a concise description.
The planning module also evaluates the outcomes of actions based on feedback from the environment after each action at step $t-1$, adjusting the plan accordingly. 

The revised plan is added to the working memory which is accessed by the decision-making module, tasked with selecting the most suitable action aligned with the current plan’s objectives. This module adheres to the ReAct \cite{yao2023react} framework, requiring the agent to articulate the rationale behind an action before execution. Separation of planning from decision-making enables LLMs to focus on distinct cognitive processes. 
In text-based environments an agent selects an action from the list of valid actions. Our agent can also use graph specific function for navigation utilizing its memory module. It extends its action space with ``go to location'' type commands and infers an optimal route to a target location using spatial relations stored in a semantic graph.

\section{Experimental Setup}
\subsection{TextWorld interactive environments}

\begin{figure*}[t]
  \centering
  \makebox[\textwidth]{\includegraphics[width=.9\textwidth]{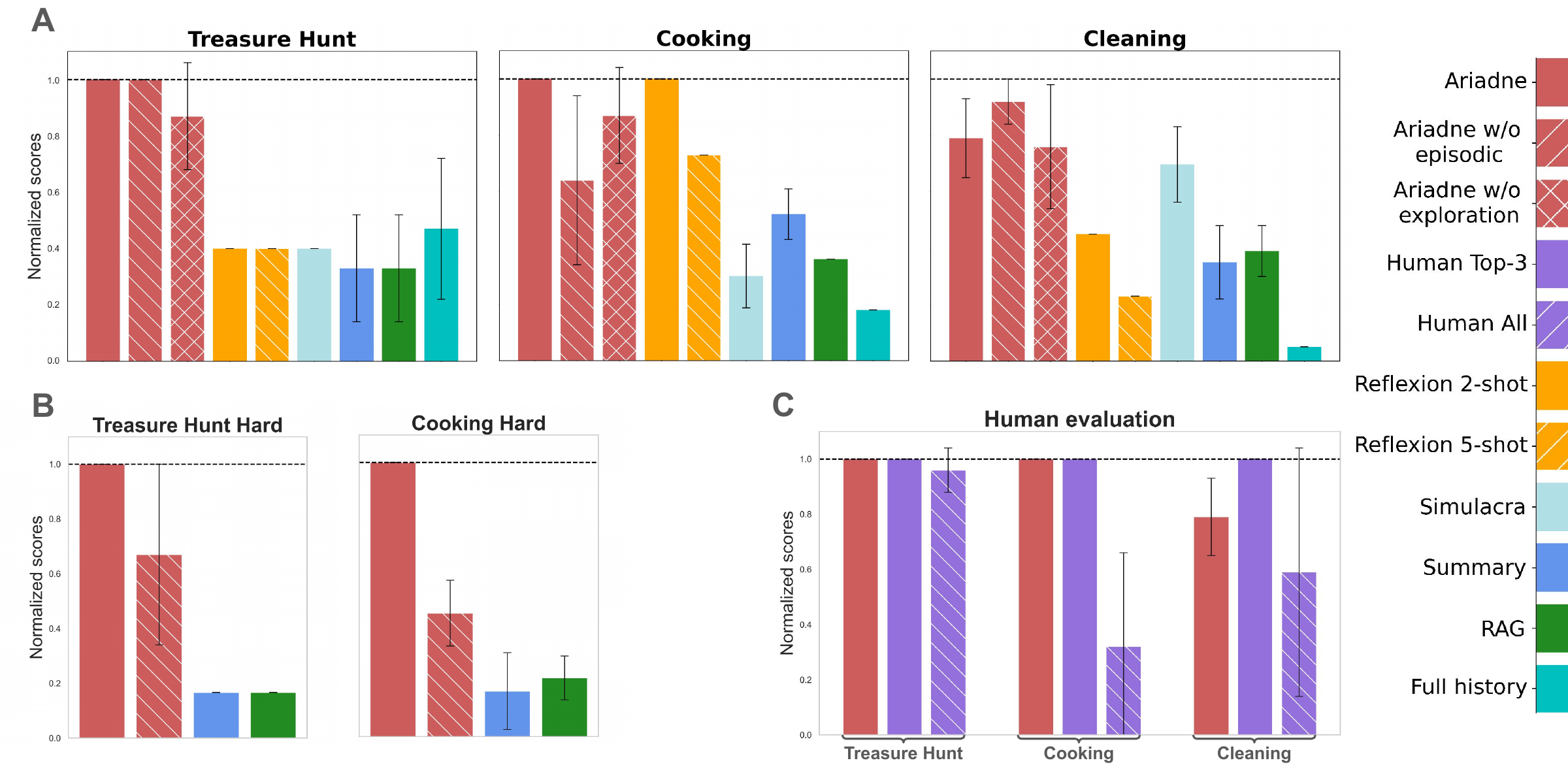}}
  \caption{AriGraph world model enables Ariadne agent to successfully solve variety of text games.
(A) Ariadne outperform baseline agents with alternative types of memory.
(B) Ariadne with episodic and semantic memory scales to harder environments without losing performance. 
  (C) Ariadne shows performance comparable to the best human players.
    The Y-axis shows the normalized score, which is calculated relative to the maximum possible points that can be obtained in each environment. Error bars show standard deviation. 
  The number of max steps is set to 60 in the Cooking and to 150 in other games. 
  }
  \label{fig:textworld_columns}
\end{figure*} 

We compared Ariadne agent with alternative methods in a series of text based games involving spatial navigation, object collection and tool manipulation. Detailed descriptions of each game, including their difficulty levels and environmental maps, can be found in the Appendix \ref{app:games}. All these games can be considered Partially Observable MDPs (POMDPs). Such games have long been benchmarks for researching agents capable of effectively remembering information and establishing long-term dependencies \cite{parisotto2020stabilizing,pleines2022memory,sorokin2022explain}. 

\textbf{Treasure Hunting.} The primary objective is to retrieve the hidden treasure, with a series of rooms providing keys and clues leading to the final goal. The basic variation has 12 rooms and 4 keys, hard one has 16 rooms and 5 keys and hardest contains 36 rooms, 7 keys and additional distract items in every room.

\textbf{Cleaning.} The goal is to clean a house by identifying and returning misplaced items to their correct locations. Environment consists of  9 rooms (kitchen, pool, etc.) and contains 11 misplaced items (among many other items). To solve the problem, the agent needs to memorize the location of rooms and objects, as well as reason about objects placement.

\textbf{Cooking.} The goal is to prepare and consume a meal by following a recipe, selecting the correct ingredients, and using appropriate tools, while navigating in multi-room house. The task is testing agents ability to remember relevant information and plan according to it. Basic difficulty task features 9 locations and 3 ingredients and hard task features 12 locations and 4 ingredients, while hardest task also features closed doors and inventory management.

For baselines we used Ariadne's planning and decision making module with one of the following types of memory instead of AriGraph model: full history of observations and actions, iterative summarization, RAG, RAG with Reflexion \cite{shinn2023reflexion}, and Simulacra - memory implementation from \cite{park2023generative}.

\textit{Full history} involves retaining a complete record of all observations and actions to inform decision-making at every step. 
\textit{Summarization}, as an alternative to storing the full history, focuses on retaining only the necessary information while discarding the rest. 
The standard \textit{RAG} baseline retrieves top-k memories based on their similarity score to the current observation and plan. \textit{Simulacra} features a scoring mechanism that integrates recency, importance, and relevance, alongside reflections on the extracted memories. 
The \textit{Reflexion} baseline differs from other methods in its approach, as it operates over multiple trials. After failing a trial, the agent reflects on its trajectories to document information that may assist in solving the task in subsequent trials.
We used the gpt-4-0125-preview as LLM backbone for AriGraph and other LLM-based baselines. 

Additionally, we tested our architecture on a variation of the cooking test from \cite{adhikari2021learning} to compare it with RL baselines. These tasks have 4 levels of difficulty, however, they are significantly simpler than our main tasks, having fewer locations, ingredients, and required actions (Appendix \ref{app:games}).

For \textit{RL} baselines, we collect the best results reported by \cite{adhikari2021learning,tuli2022learning,basu2024explorer} for the GATA, LTL-GATA, and EXPLORER architectures on the Cooking task with four difficulties levels from \cite{adhikari2021learning}.

To estimate human performance in the same games, we developed a graphical user interface 
, allowing volunteers to play basic versions of the Treasure Hunt, The Cleaning, and the Cooking. After collecting the data, we excluded sessions where the game was not completed.

\subsection{NetHack environment}

NetHack~\cite{NethackLE_kuttler2020} is a classic roguelike adventure game featuring procedurally generated multi-level dungeon (see Figure \ref{fig:nethack_env} in Appendix for a dungeon level example). It poses significant challenges for both LLM-based and RL-based approaches, requiring complex exploration, resource management, and strategic planning. 
 
We based our experiments on NetPlay~\cite{netplay_jeurissen2024} agent, which demonstrates state-of-the-art performance among LLM agents that do not rely on finetuning or RL. In NetPlay agent receives textual observations containing all information about current explored dungeon level. These observations (\emph{Level obs}) effectively function as handcrafted memory oracle for the agent. 

To evaluate our Ariadne agent, we restricted textual observations to agent's current room or corridor (\emph{Room Obs}), testing whether AriGraph world model could compensate for this restriction by remembering all relevant level information.

We compare three agents. The first is \emph{NetPlay [Room obs]} with restricted textual observations, the second is our \emph{Ariadne [Room obs]} agent that receives Room Obs and updates AriGraph, and the last is \emph{NetPlay [Level obs]} with access to inforation about explored level.

\subsection{Multi-hop Q\&A}
Although our memory architecture was originally designed for an agent interacting with the environment, we evaluated its performance on standard multi-hop Q\&A benchmarks — Musique \cite{trivedi2021musique} and HotpotQA \cite{yang2018hotpotqadatasetdiverseexplainable} to show its robustness and efficiency in more standard retrieval tasks. We made slight adjustments to the promts and replaced Contriever model with BGE-M3\cite{chen2024bgem3embeddingmultilingualmultifunctionality}, as it is a better fit for general text encoding. We used 200 random samples from both datasets similar to \cite{li2024graphreaderbuildinggraphbasedagent}. 
We compared the performance of our approach against Graphreader  
\cite{li2024graphreaderbuildinggraphbasedagent}, ReadAgent \cite{lee2024humaninspiredreadingagentgist}, HOLMES \cite{panda2024holmeshyperrelationalknowledgegraphs}, GraphRAG \cite{edge2024localglobalgraphrag}  and RAG baselines provided in \cite{li2024graphreaderbuildinggraphbasedagent}.

\section{Results}
\subsection{TextWorld}

Every LLM based agent had five attempts to solve each game. The normalized score of one means that an agent completed the game, and score less than one represents intermediate progress.
Results on text-based games are shown on the Figure \ref{fig:textworld_columns} (for dynamics see  Appendix \ref{app:textworld_extra_results}). We estimate performance as average of three best runs. 
Ariadne successfully remembers and uses information about state of the world for all three tasks.
Baseline agents are unable to solve the Treasure Hunt, and fail to find even second key in the Treasure Hunt Hardest. 
On the other hand, Ariadne successfully solves the Treasure Hunt in about fifty steps, maintains robust performance in the Treasure Hunt Hard, and is able to complete the Treasure Hunt Hardest  with more then double amount of rooms compared to Hard version, additional keys and distractors (see Appendix \ref{app:textworld_extra_results}).    

Compared to the Treasure Hunt, the Cleaning game possesses a slightly different challenge as it is more important to properly filter outdated information about object locations, than not to lose any information. This is evident from the reduced usefulness of Episodic Memory in Ariadne agent and Full history baseline, since both memory modules focus on retaining long-term information. Overall Ariadne notably outperforms alternatives in this game. Moreover, Ariadne also outperforms Reflexion, which has additional information between episodes \cite{shinn2023reflexion}. This baseline shows markable performance growth (in comparison wuth RAG) at the second try, but degrades with following tries.

The Cooking game has the highest difficulty, because any error at intermediate step prevents completion of the whole game. All baseline agents (except Reflexion 2-shot with obvious advantage over other methods) fail to complete cooking tasks due to insufficient or misused information. In this game, episodic memory is particularly important, allowing the agent to recall useful observations such as the content of the recipe or cooking instructions. For token usage of every method see Table \ref{tab:token_usage}, Appendix \ref{app:toke_usage}. 

\begin{figure}[t]
  \centering
  \includegraphics[width=0.49\textwidth]{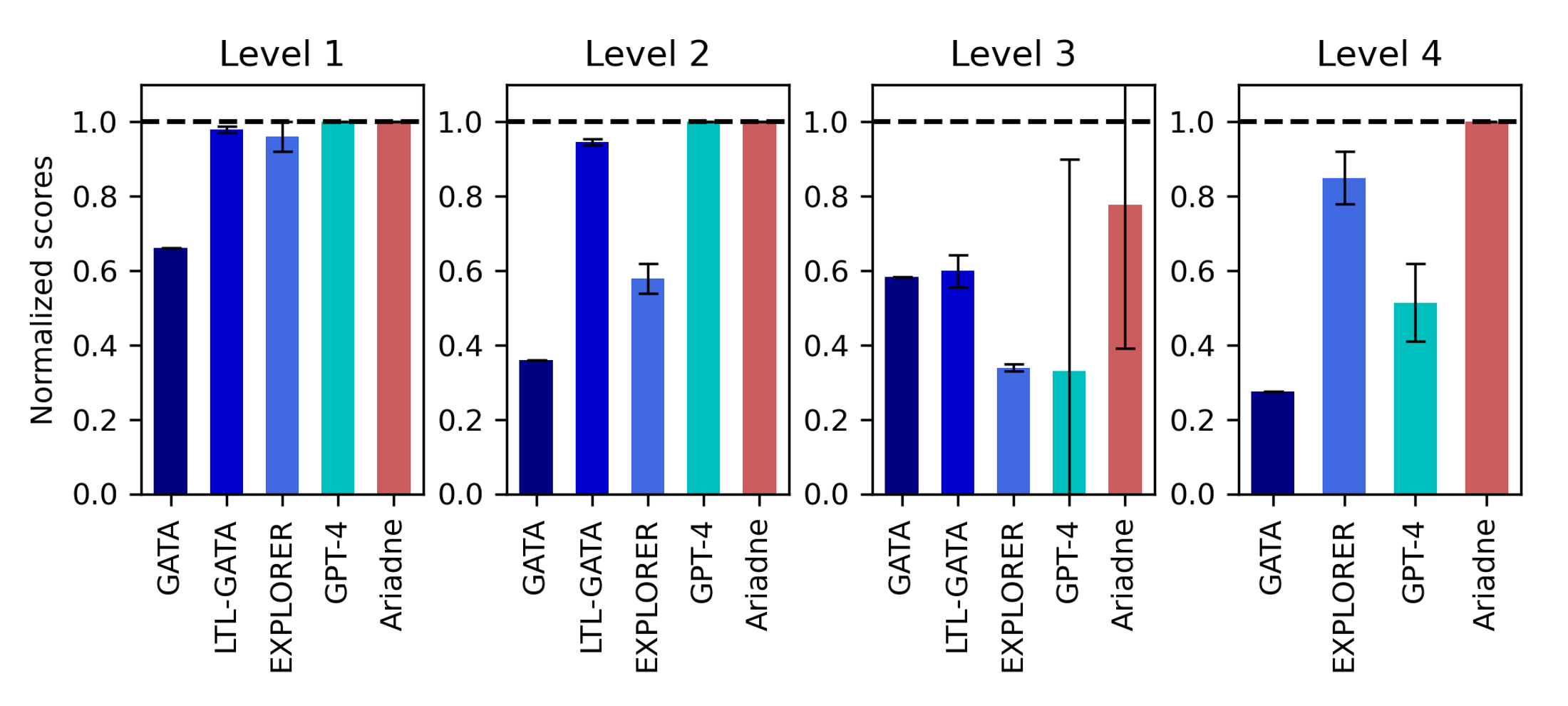}
  \caption{Ariadne LLM agent shows top performance compared to RL alternatives. Comparison of Ariadne and Full History baseline (GPT-4) with RL baselines in the cooking benchmark. Ariadne demonstrates superior performance across all 4 difficulty levels}   
  
  \label{fig:figColumns}
\end{figure}

Comparison with RL baselines on variation of the Cooking task is shown in Figure \ref{fig:figColumns}. We run Ariadne and GPT-4 with Full history on 4 difficulty levels from the cooking benchmark ~\cite{adhikari2021learning}. 
Ariadne shows superior performance to RL-agents on all 4 levels, especially harder ones. GPT-4 agent with Full history solves only first two levels which is consistent with previous result as the Cooking from Figure \ref{fig:textworld_columns}.A is harder than level 4.

\textbf{Human evaluation.}  Comparison with respect to human players is shown in Figure \ref{fig:textworld_columns}.C. \emph{All Humans} is the average score of all valid (completed) human trials. \emph{Human Top-3} is the average score of three best plays for each task. Ariadne outperforms average human player from our sample on all tasks, and scores similarly to the best human plays in the Cooking and the Treasure Hunt, but underperforms in the Cleaning.

\begin{figure*}[t]
  \center{\includegraphics[width=2\columnwidth]{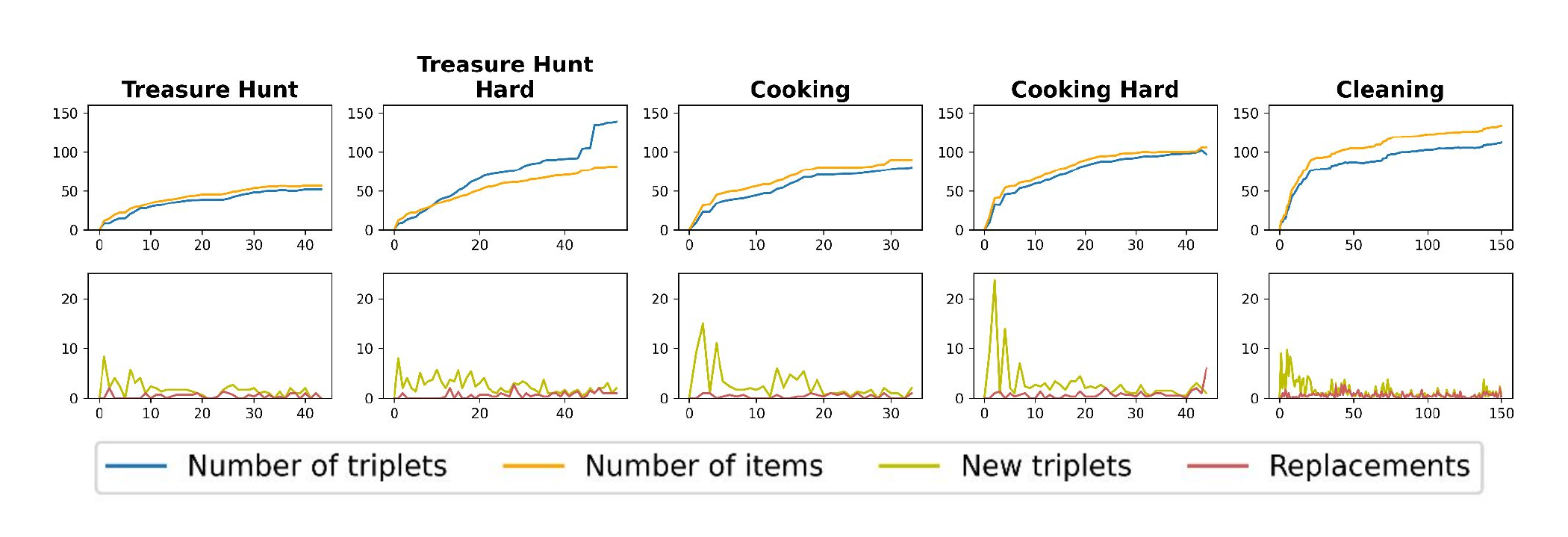}}
  \caption{AriGraph demonstrate good scaling during learning and with environment size. A size of the knowledge graph quickly saturates during exploration and learning phase. KG grows moderately when the Treasure Hunt and the Cooking games include more rooms and objects in their hard versions.}
  
  \label{fig:graphstats}
\end{figure*}

\textbf{Graph quality.} 
We measured AriGraph's growth rate and update rate during gameplay (see Figure \ref{fig:graphstats}). 
The graph actively grows during the exploration phase and flattens once the agent becomes familiar with the environment. We argue that this indicates that agent can generalize to long interactions with the environment despite constant updates to the semantic graph.
Additional results in Appendix \ref{app:graphstats} demonstrate that the growth rate of the graph decreases with the increase in quality of LLM backbone.

Overal results demonstrate clear advantage of Ariadne agent over LLM based and RL baselines.
Semantic memory enables the Ariadne Agent to build and update knowledge about the current state of the POMDP environment, which is crucial for navigation, exploration and capturing relevant details in interactive environments. On the other hand, episodic memory assists the agent in retrieving detailed long-term information that may not be captured in semantic memory, as demonstrated by the results in the Cooking task.

\subsection{NetHack}
The results are presented in Table \ref{table:nethack}. 
Scores column shows average game score across 3 runs, Levels column shows average number of dungeon levels completed by an agent. GPT-4o was used for all agents. Underscoring the importance of memory in this task, \emph{NetPlay [Level obs]} with access to memory oracle  achieved the highest scores, while \emph{NetPlay [Room obs]} with only current room observations performed the worst. \emph{Ariadne [Room obs]} successfully utilized AriGraph word model, achieving performance comparable to the baseline with memory oracle.

\begin{table}[h!]
\centering
\caption{Ariadne with obscured partial observations performs comparable to NetPlay agent full level information.}
\label{table:nethack}
\begin{tabular}{lcc}
\toprule
\textbf{Method} & \textbf{Score} & \textbf{Levels} \\ 
\midrule
Ariadne (Room obs)   & $593.00 \pm 202.62$      & $6.33 \pm 2.31$                     \\ 
NetPlay (Room obs)   & $341.67 \pm 109.14$      & $3.67 \pm 1.15$                     \\ 
NetPlay (Level obs)      & $675.33 \pm 130.27$      & $7.33 \pm 1.15$                     \\ 
\bottomrule
\end{tabular}
\end{table}

\subsection{Multi-hop Q\&A}

We compared AriGraph with the latest LLM-based approaches that employ knowledge graph construction and retrieval techniques for question answering over documents (Table \ref{qa_res_table}). Our memory architecture, adapted from the Ariadne TextWorld agent, utilizing both GPT-4 and GPT-4o-mini outperformed baseline methods like ReadAgent (GPT-4), GPT-4 RAG, GPT-4 full context and GraphReader (GPT-4). GraphRAG served as a strong GPT-4o-mini baseline, due to its extremely hight costs. ArigGraph (GPT-4o-mini) showed weaker performance on Musique, but outperformed GraphRAG on HotpotQA. Notably, our approach is more then 10x cheaper in comparison to GraphRAG (Table \ref{tab:token_usage}, Appendix \ref{app:toke_usage}). 

The best performance using GPT-4 was achieved by HOLMES, but AriGraph (GPT-4) exhibited comparable results. Notably, all baseline methods were specifically designed for Q\&A tasks, incorporating task-specific prompt tuning and additional architectural enhancements. Both GraphRAG and HOLMES employ hyper-relations in their graphs to connect source data with extracted entities, similar to our method. However, these approaches lack mechanisms for updates in dynamic environments, a key advantage of AriGraph.

\begin{table}[h]
    \renewcommand{\arraystretch}{1.0} 
    \setlength{\tabcolsep}{0.2cm} 
    \centering
    \caption{AriGraph memory demonstrates competitive performance on Multi-Hop Q\&A datasets. Even in non interactive tasks AriGraph is comparable to strong QA baseline agents. The best results with the base GPT-4o and GPT-4o-mini are shown in bold and underline respectively.}
    \begin{tabular}{p{3.7cm}|p{0.6cm} p{0.6cm}|p{0.6cm} p{0.6cm}}
        \toprule
        \textbf{Method} & \multicolumn{2}{c|}{ \textbf{MuSiQue}} & \multicolumn{2}{c}{\textbf{HotpotQA}} \\
        \cmidrule(r){2-3} \cmidrule(r){4-5}
        & EM & F1 & EM & F1 \\
        \midrule
        BM25(top-3) & 25.0 & 31.1 & 45.7 & 58.5 \\
        Ada-002(top-3) & 24.5 & 32.1 & 45.0 & 58.1 \\
        \midrule
        GPT-4 full context & 33.5 & 42.7 & 53.0 & 68.4 \\
        GPT-4 + supporting facts & 45.0 & 56.0 & 57.0 & 73.8 \\
        \midrule
        ReadAgent(GPT-4) & 35.0 & 45.1 & 48.0 & 62.0 \\
        GraphReader(GPT-4) & 38.0 & 47.4 & 55.0 & 70.0 \\
        HOLMES(GPT-4) & \textbf{48.0} & \textbf{58.0} & 66.0 & \textbf{78.0} \\
        AriGraph(GPT-4) & 45.0 & 57.0 & \textbf{68.0} & 74.7 \\
        \midrule
        GraphRAG(GPT-4o-mini) & \underline{40.0} & \underline{53.5} & 58.7 & 63.3 \\
        AriGraph(GPT-4o-mini) & 36.5 & 47.9 & \underline{60.0} & \underline{68.6} \\
        \bottomrule
    \end{tabular}
    
    \label{qa_res_table}
\end{table}

\section{Related work}

Voyager \cite{wang2023voyager}, Ghost in the Minecraft \cite{zhu2023ghost} and Jarvis-1 \cite{wang2023jarvis1} are advanced, open-ended LLM agents that show significantly better performance in Minecraft compared to earlier techniques. These agents feature memory capabilities through a library of learned skills, summaries of successful actions, and episodic memory with plans for successful task execution. However, they fall short in representing knowledge with semantic structure and depend heavily on the LLM's extensive Minecraft knowledge, or even access to the Minecraft wiki.
Generative agents \cite{park2023generative} mimic human behavior in multi-agent environments and were among the pioneers in introducing an advanced memory system for LLM agents, which we use as our baseline. 
Reflexion \cite{shinn2023reflexion} and CLIN \cite{majumder2023clincontinuallylearninglanguage} enables agents to reflect on past trajectories, allowing them to store relevant insights about completed actions in a long-term memory module, but has no structural representation of knowledge and episodic memories.
LARP \cite{yan2023larp} utilizes the concepts of episodic and semantic memories but treats them as separate instances and lacks a structural representation of knowledge.

Considerable research is dedicated to leveraging established knowledge graphs for enhancing Q\&A \cite{baek2023knowledgeaugmented,
li2024enhanced} systems  to address the factual knowledge deficiency observed in LLMs. 
The latest research demonstrating best performance in Q\&A tasks includes Graphreader  
\cite{li2024graphreaderbuildinggraphbasedagent}, HOLMES \cite{panda2024holmeshyperrelationalknowledgegraphs}, HippoRAG \cite{gutiérrez2024hipporagneurobiologicallyinspiredlongterm}, GraphRAG \cite{edge2024localglobalgraphrag} which all employ the technique of building knowledge graphs from texts. However, these studies do not address the context of functioning within an interactive environment, nor do they take into account the updates to knowledge graphs prompted by new experiences. 

Text-based environments \cite{cote18textworld,hausknecht19,ALFWorld20,wang2022scienceworld} were originally designed to evaluate reinforcement learning (RL) agents \cite{guo2020interactive,yao2020calm,ammanabrolu2020avoid,ammanabrolu2020graph,tuli2022learning,adhikari2021learning}.
Multiple experiments have already explored the potential of LLMs in these complex scenarios \cite{tsai2023large,tan2023textbased,momennejad2023evaluating,ding2024mango}. However raw LLMs show poor results in these games without proper agentic architecture and memory.

\section{Conclusions}

In this paper, we introduced AriGraph, a novel knowledge graph world model tailored for LLM agents. AriGraph uniquely integrates semantic and episodic memories from textual observations, providing a structured and dynamic representation of knowledge. We evaluated this approach across a range of interactive text-based games and multi-hop Q\&A benchmarks, comparing it against existing memory architectures. To test its capabilities comprehensively, we developed a cognitive architecture called Ariadne, which combines AriGraph with planning and decision-making components.

Our results demonstrate that AriGraph significantly outperforms other memory systems in tasks requiring long-term memory, such as decision-making, planning, and exploration in partially observable environments. The structured knowledge representation provided by AriGraph enables efficient retrieval and reasoning, accelerating learning and task completion. Additionally, AriGraph’s scalability was evident as it maintained high performance even when the complexity of tasks increased, involving more objects and locations. In multi-hop Q\&A benchmarks, AriGraph exhibited competitive performance, underscoring its robustness and adaptability beyond interactive environments.

While promising, our approach can be further enhanced by incorporating multi-modal observations, procedural memories, and more sophisticated graph search methods.

\clearpage
\bibliographystyle{named}
\bibliography{ijcai25}

\begin{thebibliography}{}

\bibitem[\protect\citeauthoryear{Adhikari \bgroup \em et al.\egroup }{2021}]{adhikari2021learning}
Ashutosh Adhikari, Xingdi Yuan, Marc-Alexandre Côté, Mikuláš Zelinka, Marc-Antoine Rondeau, Romain Laroche, Pascal Poupart, Jian Tang, Adam Trischler, and William~L. Hamilton.
\newblock Learning dynamic belief graphs to generalize on text-based games, 2021.

\bibitem[\protect\citeauthoryear{Ammanabrolu and Hausknecht}{2020}]{ammanabrolu2020graph}
Prithviraj Ammanabrolu and Matthew Hausknecht.
\newblock Graph constrained reinforcement learning for natural language action spaces, 2020.

\bibitem[\protect\citeauthoryear{Ammanabrolu \bgroup \em et al.\egroup }{2020}]{ammanabrolu2020avoid}
Prithviraj Ammanabrolu, Ethan Tien, Matthew Hausknecht, and Mark~O. Riedl.
\newblock How to avoid being eaten by a grue: Structured exploration strategies for textual worlds, 2020.

\bibitem[\protect\citeauthoryear{Baek \bgroup \em et al.\egroup }{2023}]{baek2023knowledgeaugmented}
Jinheon Baek, Alham~Fikri Aji, and Amir Saffari.
\newblock Knowledge-augmented language model prompting for zero-shot knowledge graph question answering, 2023.

\bibitem[\protect\citeauthoryear{Basu \bgroup \em et al.\egroup }{2024}]{basu2024explorer}
Kinjal Basu, Keerthiram Murugesan, Subhajit Chaudhury, Murray Campbell, Kartik Talamadupula, and Tim Klinger.
\newblock Explorer: Exploration-guided reasoning for textual reinforcement learning, 2024.

\bibitem[\protect\citeauthoryear{Bulatov \bgroup \em et al.\egroup }{2022}]{bulatov2022recurrent}
Aydar Bulatov, Yuri Kuratov, and Mikhail~S. Burtsev.
\newblock Recurrent memory transformer, 2022.

\bibitem[\protect\citeauthoryear{Bulatov \bgroup \em et al.\egroup }{2024}]{bulatov2024scaling}
Aydar Bulatov, Yuri Kuratov, Yermek Kapushev, and Mikhail~S. Burtsev.
\newblock Scaling transformer to 1m tokens and beyond with rmt, 2024.

\bibitem[\protect\citeauthoryear{Chen \bgroup \em et al.\egroup }{2024}]{chen2024bgem3embeddingmultilingualmultifunctionality}
Jianlv Chen, Shitao Xiao, Peitian Zhang, Kun Luo, Defu Lian, and Zheng Liu.
\newblock Bge m3-embedding: Multi-lingual, multi-functionality, multi-granularity text embeddings through self-knowledge distillation, 2024.

\bibitem[\protect\citeauthoryear{Cheng \bgroup \em et al.\egroup }{2024}]{cheng2024exploring}
Yuheng Cheng, Ceyao Zhang, Zhengwen Zhang, Xiangrui Meng, Sirui Hong, Wenhao Li, Zihao Wang, Zekai Wang, Feng Yin, Junhua Zhao, and Xiuqiang He.
\newblock Exploring large language model based intelligent agents: Definitions, methods, and prospects, 2024.

\bibitem[\protect\citeauthoryear{C\^ot\'e \bgroup \em et al.\egroup }{2018}]{cote18textworld}
Marc-Alexandre C\^ot\'e, \'Akos K\'ad\'ar, Xingdi Yuan, Ben Kybartas, Tavian Barnes, Emery Fine, James Moore, Ruo~Yu Tao, Matthew Hausknecht, Layla~El Asri, Mahmoud Adada, Wendy Tay, and Adam Trischler.
\newblock Textworld: A learning environment for text-based games.
\newblock {\em CoRR}, abs/1806.11532, 2018.

\bibitem[\protect\citeauthoryear{Ding \bgroup \em et al.\egroup }{2024a}]{ding2024mango}
Peng Ding, Jiading Fang, Peng Li, Kangrui Wang, Xiaochen Zhou, Mo~Yu, Jing Li, Matthew Walter, and Hongyuan Mei.
\newblock {MANGO}: A benchmark for evaluating mapping and navigation abilities of large language models, 2024.

\bibitem[\protect\citeauthoryear{Ding \bgroup \em et al.\egroup }{2024b}]{ding2024longrope}
Yiran Ding, Li~Lyna Zhang, Chengruidong Zhang, Yuanyuan Xu, Ning Shang, Jiahang Xu, Fan Yang, and Mao Yang.
\newblock Longrope: Extending llm context window beyond 2 million tokens, 2024.

\bibitem[\protect\citeauthoryear{Edge \bgroup \em et al.\egroup }{2024}]{edge2024localglobalgraphrag}
Darren Edge, Ha~Trinh, Newman Cheng, Joshua Bradley, Alex Chao, Apurva Mody, Steven Truitt, and Jonathan Larson.
\newblock From local to global: A graph rag approach to query-focused summarization, 2024.

\bibitem[\protect\citeauthoryear{Gu and Dao}{2023}]{gu2023mamba}
Albert Gu and Tri Dao.
\newblock Mamba: Linear-time sequence modeling with selective state spaces, 2023.

\bibitem[\protect\citeauthoryear{Guo \bgroup \em et al.\egroup }{2020}]{guo2020interactive}
Xiaoxiao Guo, Mo~Yu, Yupeng Gao, Chuang Gan, Murray Campbell, and Shiyu Chang.
\newblock Interactive fiction game playing as multi-paragraph reading comprehension with reinforcement learning, 2020.

\bibitem[\protect\citeauthoryear{Gutiérrez \bgroup \em et al.\egroup }{2024}]{gutiérrez2024hipporagneurobiologicallyinspiredlongterm}
Bernal~Jiménez Gutiérrez, Yiheng Shu, Yu~Gu, Michihiro Yasunaga, and Yu~Su.
\newblock Hipporag: Neurobiologically inspired long-term memory for large language models, 2024.

\bibitem[\protect\citeauthoryear{Hausknecht \bgroup \em et al.\egroup }{2019}]{hausknecht19}
Matthew Hausknecht, Prithviraj Ammanabrolu, C\^ot\'{e} Marc-Alexandre, and Yuan Xingdi.
\newblock Interactive fiction games: A colossal adventure.
\newblock {\em CoRR}, abs/1909.05398, 2019.

\bibitem[\protect\citeauthoryear{Izacard \bgroup \em et al.\egroup }{2022}]{izacard2022unsupervised}
Gautier Izacard, Mathilde Caron, Lucas Hosseini, Sebastian Riedel, Piotr Bojanowski, Armand Joulin, and Edouard Grave.
\newblock Unsupervised dense information retrieval with contrastive learning, 2022.

\bibitem[\protect\citeauthoryear{Jeurissen \bgroup \em et al.\egroup }{2024}]{netplay_jeurissen2024}
Dominik Jeurissen, Diego Perez-Liebana, Jeremy Gow, Duygu Cakmak, and James Kwan.
\newblock Playing nethack with llms: Potential and limitations as zero-shot agents, 2024.

\bibitem[\protect\citeauthoryear{K{\"u}ttler \bgroup \em et al.\egroup }{2020}]{NethackLE_kuttler2020}
Heinrich K{\"u}ttler, Nantas Nardelli, Alexander Miller, Roberta Raileanu, Marco Selvatici, Edward Grefenstette, and Tim Rockt{\"a}schel.
\newblock The nethack learning environment.
\newblock {\em Advances in Neural Information Processing Systems}, 33:7671--7684, 2020.

\bibitem[\protect\citeauthoryear{Lee \bgroup \em et al.\egroup }{2024}]{lee2024humaninspiredreadingagentgist}
Kuang-Huei Lee, Xinyun Chen, Hiroki Furuta, John Canny, and Ian Fischer.
\newblock A human-inspired reading agent with gist memory of very long contexts, 2024.

\bibitem[\protect\citeauthoryear{Li \bgroup \em et al.\egroup }{2024a}]{li2024graphreaderbuildinggraphbasedagent}
Shilong Li, Yancheng He, Hangyu Guo, Xingyuan Bu, Ge~Bai, Jie Liu, Jiaheng Liu, Xingwei Qu, Yangguang Li, Wanli Ouyang, Wenbo Su, and Bo~Zheng.
\newblock Graphreader: Building graph-based agent to enhance long-context abilities of large language models, 2024.

\bibitem[\protect\citeauthoryear{Li \bgroup \em et al.\egroup }{2024b}]{li2024enhanced}
Yihao Li, Ru~Zhang, Jianyi Liu, and Gongshen Liu.
\newblock An enhanced prompt-based llm reasoning scheme via knowledge graph-integrated collaboration, 2024.

\bibitem[\protect\citeauthoryear{Majumder \bgroup \em et al.\egroup }{2023}]{majumder2023clincontinuallylearninglanguage}
Bodhisattwa~Prasad Majumder, Bhavana~Dalvi Mishra, Peter Jansen, Oyvind Tafjord, Niket Tandon, Li~Zhang, Chris Callison-Burch, and Peter Clark.
\newblock Clin: A continually learning language agent for rapid task adaptation and generalization, 2023.

\bibitem[\protect\citeauthoryear{Momennejad \bgroup \em et al.\egroup }{2023}]{momennejad2023evaluating}
Ida Momennejad, Hosein Hasanbeig, Felipe~Vieira Frujeri, Hiteshi Sharma, Nebojsa Jojic, Hamid Palangi, Robert Ness, and Jonathan Larson.
\newblock Evaluating cognitive maps and planning in large language models with cogeval.
\newblock In {\em Thirty-seventh Conference on Neural Information Processing Systems}, 2023.

\bibitem[\protect\citeauthoryear{Murugesan \bgroup \em et al.\egroup }{2020}]{murugesan2020textbased}
Keerthiram Murugesan, Mattia Atzeni, Pavan Kapanipathi, Pushkar Shukla, Sadhana Kumaravel, Gerald Tesauro, Kartik Talamadupula, Mrinmaya Sachan, and Murray Campbell.
\newblock Text-based rl agents with commonsense knowledge: New challenges, environments and baselines, 2020.

\bibitem[\protect\citeauthoryear{Pan \bgroup \em et al.\egroup }{2024}]{Pan_2024}
Shirui Pan, Linhao Luo, Yufei Wang, Chen Chen, Jiapu Wang, and Xindong Wu.
\newblock Unifying large language models and knowledge graphs: A roadmap.
\newblock {\em IEEE Transactions on Knowledge and Data Engineering}, page 1–20, 2024.

\bibitem[\protect\citeauthoryear{Panda \bgroup \em et al.\egroup }{2024}]{panda2024holmeshyperrelationalknowledgegraphs}
Pranoy Panda, Ankush Agarwal, Chaitanya Devaguptapu, Manohar Kaul, and Prathosh~A P.
\newblock Holmes: Hyper-relational knowledge graphs for multi-hop question answering using llms, 2024.

\bibitem[\protect\citeauthoryear{Parisotto \bgroup \em et al.\egroup }{2020}]{parisotto2020stabilizing}
Emilio Parisotto, Francis Song, Jack Rae, Razvan Pascanu, Caglar Gulcehre, Siddhant Jayakumar, Max Jaderberg, Raphael~Lopez Kaufman, Aidan Clark, Seb Noury, et~al.
\newblock Stabilizing transformers for reinforcement learning.
\newblock In {\em International conference on machine learning}, pages 7487--7498. PMLR, 2020.

\bibitem[\protect\citeauthoryear{Park \bgroup \em et al.\egroup }{2023}]{park2023generative}
Joon~Sung Park, Joseph~C. O'Brien, Carrie~J. Cai, Meredith~Ringel Morris, Percy Liang, and Michael~S. Bernstein.
\newblock Generative agents: Interactive simulacra of human behavior, 2023.

\bibitem[\protect\citeauthoryear{Pleines \bgroup \em et al.\egroup }{2022}]{pleines2022memory}
Marco Pleines, Matthias Pallasch, Frank Zimmer, and Mike Preuss.
\newblock Memory gym: Partially observable challenges to memory-based agents.
\newblock In {\em The eleventh international conference on learning representations}, 2022.

\bibitem[\protect\citeauthoryear{Shinn \bgroup \em et al.\egroup }{2023}]{shinn2023reflexion}
Noah Shinn, Federico Cassano, Edward Berman, Ashwin Gopinath, Karthik Narasimhan, and Shunyu Yao.
\newblock Reflexion: Language agents with verbal reinforcement learning, 2023.

\bibitem[\protect\citeauthoryear{Shridhar \bgroup \em et al.\egroup }{2021}]{ALFWorld20}
Mohit Shridhar, Xingdi Yuan, Marc-Alexandre C\^ot\'e, Yonatan Bisk, Adam Trischler, and Matthew Hausknecht.
\newblock {ALFWorld: Aligning Text and Embodied Environments for Interactive Learning}.
\newblock In {\em Proceedings of the International Conference on Learning Representations (ICLR)}, 2021.

\bibitem[\protect\citeauthoryear{Sorokin \bgroup \em et al.\egroup }{2022}]{sorokin2022explain}
Artyom Sorokin, Nazar Buzun, Leonid Pugachev, and Mikhail Burtsev.
\newblock Explain my surprise: Learning efficient long-term memory by predicting uncertain outcomes.
\newblock {\em Advances in Neural Information Processing Systems}, 35:36875--36888, 2022.

\bibitem[\protect\citeauthoryear{Sumers \bgroup \em et al.\egroup }{2024}]{sumers2024cognitive}
Theodore~R. Sumers, Shunyu Yao, Karthik Narasimhan, and Thomas~L. Griffiths.
\newblock Cognitive architectures for language agents, 2024.

\bibitem[\protect\citeauthoryear{Tan \bgroup \em et al.\egroup }{2023}]{tan2023textbased}
Qinyue Tan, Ashkan Kazemi, and Rada Mihalcea.
\newblock Text-based games as a challenging benchmark for large language models, 2023.

\bibitem[\protect\citeauthoryear{Trivedi \bgroup \em et al.\egroup }{2022}]{trivedi2021musique}
Harsh Trivedi, Niranjan Balasubramanian, Tushar Khot, and Ashish Sabharwal.
\newblock {M}u{S}i{Q}ue: Multihop questions via single-hop question composition.
\newblock {\em Transactions of the Association for Computational Linguistics}, 2022.

\bibitem[\protect\citeauthoryear{Tsai \bgroup \em et al.\egroup }{2023}]{tsai2023large}
Chen~Feng Tsai, Xiaochen Zhou, Sierra~S. Liu, Jing Li, Mo~Yu, and Hongyuan Mei.
\newblock Can large language models play text games well? current state-of-the-art and open questions, 2023.

\bibitem[\protect\citeauthoryear{Tuli \bgroup \em et al.\egroup }{2022}]{tuli2022learning}
Mathieu Tuli, Andrew~C. Li, Pashootan Vaezipoor, Toryn~Q. Klassen, Scott Sanner, and Sheila~A. McIlraith.
\newblock Learning to follow instructions in text-based games, 2022.

\bibitem[\protect\citeauthoryear{Wang \bgroup \em et al.\egroup }{2022}]{wang2022scienceworld}
Ruoyao Wang, Peter Jansen, Marc-Alexandre Côté, and Prithviraj Ammanabrolu.
\newblock Scienceworld: Is your agent smarter than a 5th grader?, 2022.

\bibitem[\protect\citeauthoryear{Wang \bgroup \em et al.\egroup }{2023a}]{wang2023voyager}
Guanzhi Wang, Yuqi Xie, Yunfan Jiang, Ajay Mandlekar, Chaowei Xiao, Yuke Zhu, Linxi Fan, and Anima Anandkumar.
\newblock Voyager: An open-ended embodied agent with large language models, 2023.

\bibitem[\protect\citeauthoryear{Wang \bgroup \em et al.\egroup }{2023b}]{wang2023jarvis1}
Zihao Wang, Shaofei Cai, Anji Liu, Yonggang Jin, Jinbing Hou, Bowei Zhang, Haowei Lin, Zhaofeng He, Zilong Zheng, Yaodong Yang, Xiaojian Ma, and Yitao Liang.
\newblock Jarvis-1: Open-world multi-task agents with memory-augmented multimodal language models, 2023.

\bibitem[\protect\citeauthoryear{Wang \bgroup \em et al.\egroup }{2024}]{Wang_2024}
Lei Wang, Chen Ma, Xueyang Feng, Zeyu Zhang, Hao Yang, Jingsen Zhang, Zhiyuan Chen, Jiakai Tang, Xu~Chen, Yankai Lin, Wayne~Xin Zhao, Zhewei Wei, and Jirong Wen.
\newblock A survey on large language model based autonomous agents.
\newblock {\em Frontiers of Computer Science}, 18(6), March 2024.

\bibitem[\protect\citeauthoryear{Wong~Gonzalez}{2018}]{WongGonzalez2018}
Daniela Wong~Gonzalez.
\newblock {\em The Relationship Between Semantic and Episodic Memory: Exploring the Effect of Semantic Neighbourhood Density on Episodic Memory}.
\newblock PhD thesis, Electronic Theses and Dissertations, 2018.
\newblock Paper 7585.

\bibitem[\protect\citeauthoryear{Yan \bgroup \em et al.\egroup }{2023}]{yan2023larp}
Ming Yan, Ruihao Li, Hao Zhang, Hao Wang, Zhilan Yang, and Ji~Yan.
\newblock Larp: Language-agent role play for open-world games, 2023.

\bibitem[\protect\citeauthoryear{Yang \bgroup \em et al.\egroup }{2018}]{yang2018hotpotqadatasetdiverseexplainable}
Zhilin Yang, Peng Qi, Saizheng Zhang, Yoshua Bengio, William~W. Cohen, Ruslan Salakhutdinov, and Christopher~D. Manning.
\newblock Hotpotqa: A dataset for diverse, explainable multi-hop question answering, 2018.

\bibitem[\protect\citeauthoryear{Yao \bgroup \em et al.\egroup }{2020}]{yao2020calm}
Shunyu Yao, Rohan Rao, Matthew Hausknecht, and Karthik Narasimhan.
\newblock Keep calm and explore: Language models for action generation in text-based games, 2020.

\bibitem[\protect\citeauthoryear{Yao \bgroup \em et al.\egroup }{2023}]{yao2023react}
Shunyu Yao, Jeffrey Zhao, Dian Yu, Nan Du, Izhak Shafran, Karthik Narasimhan, and Yuan Cao.
\newblock React: Synergizing reasoning and acting in language models, 2023.

\bibitem[\protect\citeauthoryear{Zhu \bgroup \em et al.\egroup }{2023}]{zhu2023ghost}
Xizhou Zhu, Yuntao Chen, Hao Tian, Chenxin Tao, Weijie Su, Chenyu Yang, Gao Huang, Bin Li, Lewei Lu, Xiaogang Wang, Yu~Qiao, Zhaoxiang Zhang, and Jifeng Dai.
\newblock Ghost in the minecraft: Generally capable agents for open-world environments via large language models with text-based knowledge and memory, 2023.

\end{thebibliography}

\clearpage
\appendix

\section{Memory Graph Search Details}
\label{app:memory_graph_search}
Pseudo-code for $\texttt{SemanticSearch}$ function in AriGraph is listed in Algorithm \ref{app:algo:semantic_retrieval}. 
This algorithm is close to BFS search. The main difference is the use of retrieval mechanism in function $\texttt{EmbedAndRetrieve}$. 
Function $\texttt{EmbedAndRetrieve}(E, q, w)$ uses pretrained Contriever~\cite{izacard2022unsupervised} model to compute embeddings for edges $E$ and query $q$ and then returns top $w$ edges with a highest similarity score. Similarity score between edge $e$ and query $q$ is a dot product between their embeddings. 
Most of the times, when query for $\texttt{EmbedAndRetrieve}$ is a semantic vertex, it simply returns edges incident to this vertex, but also has ability to retrieve edges connected to vertices that are semantically close to the query vertex. For example, semantic graph can contain separate vertices for "grill" and "grilling" that are not directly connected, so searching for "grill"  can potentially return edge \mbox{("bbq", "used for", "grilling").}

\begin{algorithm}[h!] 

    \SetAlgoLined
    \KwIn{search query $q$, 
    $E_s$, search depth $d$, search width $w$}
    \KwResult{relevant vertices semantic $V^q_s$ and edges $E^q_s$}
    $E^q_s \leftarrow \emptyset$
    \\
    $L \leftarrow \emptyset$ \hspace{0.18\textwidth}\tcp{init empty queue of queries} 
    $\texttt{Enqueue}(L, q)$ \hspace{0.095\textwidth}\tcp{push q into queue $L$} 
    $D[q] \leftarrow 0$ \hspace{0.15\textwidth}\tcp{set search distance for $q$ to zero}
    \While{$L$ is not empty}{
        $q' \leftarrow \texttt{Dequeue}(L)$ \hspace{0.1\textwidth}\tcp{remove first element from $L$} 
        \If{$D[q'] \geq d$}{
            continue
        }
        \tcp{use Contriever model to find top $w$ triplets closest to $q'$}
        $E'_s \leftarrow \texttt{EmbedAndRetrieve}(E_s, q', w)$ \\
        \ForEach{$e_i$ in $E'_s$}{
            $V'_s \leftarrow \texttt{IncidentVertices}(e_i)$ \tcp{returns two incident semantic vertices} 
            \ForEach{$v$ in $V'_s$}{
                \If{$v$ not in $L$}{
                    $\texttt{Enqueue}(L, v)$ \\
                    $D[v] \leftarrow D[q'] + 1$
                }
            }
        }
        $E^q_s \leftarrow E^q_s \cup E'_s$
    }
    \Return{$E^q_s$}

\caption{Semantic Search}
\label{app:algo:semantic_retrieval}
\end{algorithm}

\section{Exploration} 
\label{app:exploration}

Before any planning or decision-making occurs, an auxiliary agent assesses the need for exploration based on a pre-established plan. Depending on this assessment, it either activates or deactivates exploration mode. Moreover, the graph agent extracts triplets containing information about exits, such as "kitchen, has an unexplored exit, south," and triplets detailing spatial connections between locations like "hall, east of, kitchen." Subsequently, simple graph-based methods can be employed to identify all exits from each room that the agent has detected but not yet explored. This information is then added to the working memory of the agent.
Function for finding triplets corresponding to probable unexplored actions is listed in Algorithm \ref{explore_exits}. Current implementation uses expert knowledge about what elements of the semantic graph can represent locations and exits between them.

\begin{algorithm}[h!]
     \KwIn{$V_s$, $E_s$, $V_e$, $E_e$, current location $v_l$}
     \KwResult{triplets $E^{exp}_s$ with information about unexplored exits from $v_l$}
     
     $E^{exp} \leftarrow \emptyset$\\
     $E^{out} \leftarrow \texttt{GetOutgoing}(E_s, v_l)$ \hspace{0.05\textwidth}\tcp{get semantic edges outgoing from $v_l$}
     
     \ForEach{$e$ in $E^{out}$}{
    
         \If{$\texttt{RepresentExit}(e)$}{ 
                \vspace{2px}
                $E^{exp} \leftarrow E^{exp} \cup \{e\}$
            } 
      } 
     $E^{in} \leftarrow \texttt{GetIncoming}(E_s, v_l)$ \hspace{0.05\textwidth}\tcp{get semantic edges incoming in $v_l$}
     \ForEach{$e$ in $E^{in}$}{
         \If{$\texttt{RepresentExit}(e)$}{
                \tcp{remove exits leading to explored locations}
                $E^{exp} \leftarrow E^{exp} \setminus \{\texttt{FindRelatedExit}(E^{exp}, e)\}$ 
            } 
      } 
      
     \Return{$E^{exp}$} 
\caption{Unexplored exit detection}
\label{explore_exits}
\end{algorithm}

\section{Graph statistics} 
\label{app:graphstats}

\begin{figure*}[t]
  \center{\includegraphics[width=0.8\textwidth]{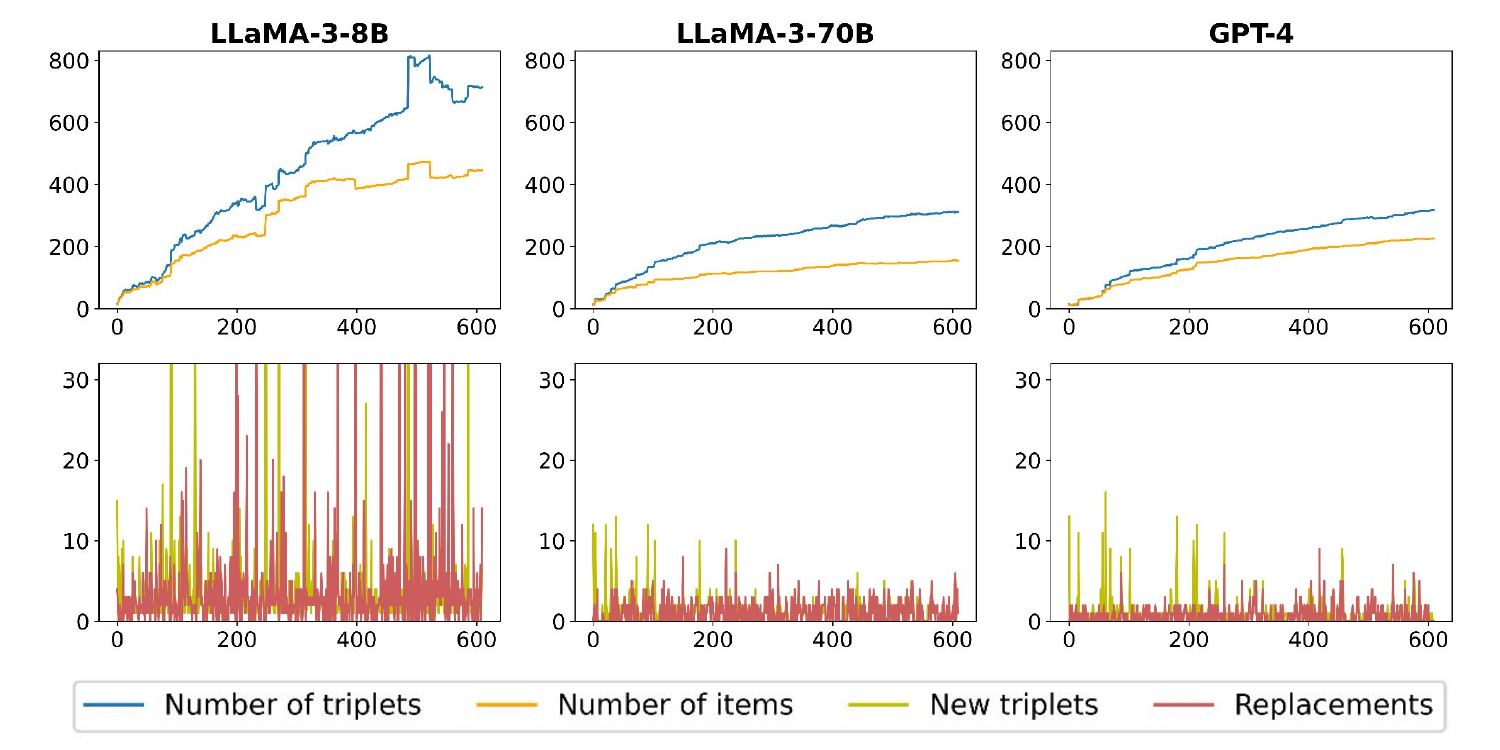}}
  \caption{\textbf{Statistics of graph construction and updating during the random walk.}}
  
  \label{app:fig:graphstats_long}
\end{figure*}
When working with a graph, it would be advantageous to assess the quality of its construction. Unfortunately, due to the variability in its construction (e.g., the use of synonyms, equivalent restructuring of connections, etc.), direct measurement of its correspondence to the true graph is extremely challenging. Moreover, the inclusion of information in the constructed graph that is not present in the true graph is not always detrimental and can sometimes yield positive effects. For these reasons, we did not measure the quality of the graph but instead measured its growth rate and update rate during gameplay (Fig.\ref{fig:graphstats}). Furthermore, we conducted a separate experiment in which the graph was constructed according to our pipeline, but the agent's actions were chosen randomly (Fig.\ref{app:fig:graphstats_long}). The Cleaning environment was chosen for this setting because, unlike the Treasure Hunt, it contains a large variety of objects with which the agent can interact, and unlike the Cooking, the agent cannot fail. In this setting, we also measured the graph's growth rate and update rate. 

\textbf{From the obtained results,} one can conclude that the graph grows most actively during the exploration phase but maintains a slower growth rate when the agent operates within the familiar parts of the environment. Additionally, it is evident that the growth rate of the graph decreases as the quality of the LLM used in its construction increases.

\section{Token usage}
\label{app:toke_usage}

\begin{table}[htbp]
  \centering
  \begin{threeparttable}
  \caption{Token Usage Analysis for Text Games and QA Tasks}
  \label{tab:token_usage}
  \begin{tabular}{@{}lrr@{}} 
    \toprule
    Method & Prompt Tokens & Completion Tokens \\
    \midrule
    \textbf{Text Games (per step)} & & \\
    Ariadne agent & 6,000 & 500 \\
    RAG memory agent & 4,000 & 350 \\
    Summary memory agent & 3,800 & 350 \\
    Full history memory agent (step 150) & 14,000 & 350 \\
    Simulacra memory agent & 7,500 & 400 \\
    Reflexion memory agent & 5,800 & 350 \\
    \midrule
    \textbf{QA Benchmarks (per task)} & & \\
    AriGraph & 11,000 & 2,500 \\
    GraphRAG & 115,000 & 20,000 \\
    \bottomrule
  \end{tabular}

  \end{threeparttable}
\end{table}

\clearpage
\section{LLM Prompts}
\label{app:prompts}

\begin{tcolorbox}[title=Triplet Extraction Prompt, enhanced jigsaw, breakable, colback=white, colframe=black, sharp corners]
\lstset{
  basicstyle=\tiny\ttfamily,
  breaklines=true,
  columns=fullflexible,
}
\begin{lstlisting}
Guidelines for Building the Knowledge Graph:

Creating Nodes and Triplets: Nodes should depict entities or concepts, similar to Wikipedia nodes. Use a structured triplet format to capture data, as follows: "subject, relation, object". For example, from "Albert Einstein, born in Germany, is known for developing the theory of relativity," extract "Albert Einstein, country of birth, Germany; Albert Einstein, developed, Theory of Relativity." 
Remember that you should break complex triplets like "John, position, engineer in Google" into simple triplets like "John, position, engineer", "John, work at, Google".
Length of your triplet should not be more than 7 words. You should extract only concrete knowledges, any assumptions must be described as hypothesis.
For example, from phrase "John have scored many points and potentiallyy will be winner" you should extract "John, scored many, points; John, could be, winner" and should not extract "John, will be, winner".
Remember that object and subject must be an atomary units while relation can be more complex and long.
If observation states that you take item, the triplet shoud be: 'item, is in, inventory' and nothing else. 

Do not miss important information. If observation is 'book involves story about knight, who needs to kill a dragon', triplets should be 'book, involves, knight', 'knight, needs to kill, dragon'. If observation involves some type of notes, do not forget to include triplets about entities this note includes.
There could be connections between distinct parts of observations. For example if there is information in the beginning of the observation that you are in location, and in the end it states that there is an exit to the east, you should extract triplet: 'location, has exit, east'. 
Several triplets can be extracted, that contain information about the same node. For example 'kitchen, contains, apple', 'kitchen, contains, table', 'apple, is on, table'. Do not miss this type of connections.
Other examples of triplets: 'room z, contains, black locker'; 'room x, has exit, east', 'apple, is on, table', 'key, is in, locker', 'apple, to be, grilled', 'potato, to be, sliced', 'stove, used for, frying', 'recipe, requires, green apple', 'recipe, requires, potato'.
Do not include triplets that state the current location of an agent like 'you, are in, location'.
Do not use 'none' as one of the entities.
If there is information that you read something, do not forget to incluse triplets that state that entitie that you read contains information that you extract.

Example of triplets you have extracted before: {example}

Observation: {observation}

Remember that triplets must be extracted in format: "subject_1, relation_1, object_1; subject_2, relation_2, object_2; ..."

Extracted triplets:'''

\end{lstlisting}
\end{tcolorbox}

\begin{tcolorbox}[title=Outdated Triplet Selection Prompt, enhanced jigsaw, breakable, colback=white, colframe=black, sharp corners]
\lstset{
  basicstyle=\tiny\ttfamily,
  breaklines=true,
  columns=fullflexible,
}
\begin{lstlisting}
The triplets denote facts about the environment where the player moves. The player takes actions and the environment changes, so some triplets from the list of existing triplets can be replaced with one of the new triplets. For example, the player took the item from the locker and the existing triplet "item, is in, locker" should be replaced with the new triplet "item, is in, inventory".

Sometimes there are no triplets to replace:
Example of existing triplets: "Golden locker, state, open"; "Room K, is west of, Room I"; "Room K, has exit, east".
Example of new triplets: "Room T, is north of, Room N"; "Room T, has exit, south".
Example of replacing: []. Nothisg to replace here

Sometimes several triplets can be replaced with one:
Example of existing triplets: "kitchen, contains, broom"; "broom, is on, floor".
Example of new triplets: "broom, is in, inventory".
Example of replacing: [["kitchen, contains, broom" -> "broom, is in, inventory"], ["broom, is on, floor" -> "broom, is in, inventory"]]. Because broom changed location from the floor in the kitchen to players inventory.

Ensure that triplets are only replaced if they contain redundant or conflicting information about the same aspect of an entity. Triplets should not be replaced if they provide distinct or complementary information about entities compared to the new triplets. Specifically, consider the relationships, properties, or contexts described by each triplet and verify that they align before replacement. If there is uncertainty about whether a triplet should be replaced, prioritize retaining the existing triplet over replacing it. When comparing existing and new triplets, if they refer to different aspects or attributes of entities, do not replace them. Replacements should only occur when there is semantic duplication between an existing triplet and a new triplet.
Example of existing triplets: "apple, to be, cooked", 'knife, used for, cutting', 'apple, has been, sliced'
Example of new triplets: "apple, is on, table", 'kitchen, contsins, knife', 'apple, has beed, grilled'.
Example of replacing: []. Nothing to replace here. These triplets describe different properties of items, so they should not be replaced. 

Another example of when not to replase existung triplets:
Example of existing triplets: "brush, used for, painting".
Example of new triplets: "brush, is in, art class".
Example of replacing: []. Nothing to replace here. These triplets describe different properties of brush, so they should not be replaced. 

I repeat, do not replace triplets, if they carry differend type of information about entities!!! It is better to leave a tripplet, than to replace the one that has important information. Do not state that triplet needs to be replaced if you are not sure!!!
If you find triplet in Existing triplets which semantically duplicate some triplet in New triplets, replace such triplet from Existing triplets. However do not replace triplets if they refer to different things. 
####
Generate only replacing, no descriptions are needed.
Existing triplets: {ex_triplets}.
New triplets: {new_triplets}.
####
Warning! Replacing must be generated strictly in following format: [[outdated_triplet_1 -> actual_triplet_1], [outdated_triplet_2 -> actual_triplet_2], ...], you MUST NOT include any descriptions in answer.
Replacing:
\end{lstlisting}
\end{tcolorbox}

\begin{tcolorbox}[title=Exploration check prompt, enhanced jigsaw, breakable, colback=white, colframe=black, sharp corners]
\lstset{
  basicstyle=\tiny\ttfamily,
  breaklines=true,
  columns=fullflexible,
}
\begin{lstlisting}
####
INSTRUCTION:
You will be provided with sub-goals and reasons for it from plan of an agent. Your task is to state if this sub goals require exploration of the environment, finding or locating something.
Answer with just True or False.
####
Plan: 
{plan0}
\end{lstlisting}
\end{tcolorbox}

\begin{tcolorbox}[title=Planning prompt, enhanced jigsaw, breakable, colback=white, colframe=black, sharp corners]
\lstset{
  basicstyle=\tiny\ttfamily,
  breaklines=true,
  columns=fullflexible,
}
\begin{lstlisting}
####
INSTRUCTION:
You are a planner within the agent system tasked with navigating the environment in a text-based game. 
Your role is to create a concise plan to achieve your main goal or modify your current plan based on new information received. 
Make sure your sub-goals will benefit the achivment of your main goal. If your main goal is an ongoing complex process, also put sub-goals that can immediately benifit achiving something from your main goal.
If you need to find something, put it into sub-goal.
If you wish to alter or delete a sub-goal within the current plan, confirm that this sub-goal has been achieved according to the current observation or is no longer relevant to achieving your main goal. Untill then do not change wording in "sub_goal" elements of your plan and their position in the plan. Only change wording in "reason" part to track the progress of completion of sub-goals.
If sub-goal was completed or confirmed to be no more relevant, delete it, replase it with new one or with lower priority sub-goals from the plan. Untill then keep the structure of sub-goals as it is. Create new sub-goals only if they will benifit your main goal and do not prioritize them over current sub-goals. 
If your task is to obtain something, make shure that the item is in your inventory before changing your sub-goal.
Your plan contains important information and goals you need to complete. Do not alter sub-goals or move them in hierarchy if they were not completed!
Pay attention to your inventory, what items you are carring, when setting the sub-goals. These items might be important.
Pay attention to information from your memory module, it is important.
There should always be at least one sub-goal.
State the progress of completing your sub-goals in "reason" for each sub-goal.

Write your answer exactly in this json format:
{ "main_goal": "...",
  "plan_steps": [{
      "sub_goal_1": "...",
      "reason": "..."
    },
    {
      "sub_goal_2": "...",
      "reason": "..."
    },
    {
      "sub_goal_...": "...",
      "reason": "..."
    }],
  "your_emotion":
    {
      "your_current_emotion": "emotion",
      "reason_behind_emotion": "..."
    }}

Do not write anything else.
####
1. Main goal: {main_goal}
2. History of {n_prev} last observations and actions: {observations} 
3. Your current observation: {observation}
4. Information from the memory module that can be relevant to current situation: {subgraph}
5. Your {topk_episodic} most relevant episodic memories from the past for the current situation: {top_episodic}.
6. Your previous plan: {plan0}
*if is explore* 7. Yet unexplored exits in the environment: {all_unexpl_exits}
\end{lstlisting}
\end{tcolorbox}

\begin{tcolorbox}[title=ReAct decision making prompt, enhanced jigsaw, breakable, colback=white, colframe=black, sharp corners]
\lstset{
  basicstyle=\tiny\ttfamily,
  breaklines=true,
  columns=fullflexible,
}
\begin{lstlisting}
####
INSTRUCTION:
You are an action selector within an agent system designed to navigate an environment in a text-based game. Your role involves receiving information about an agent and the state of the environment alongside a list of possible actions.
Your primary objective is to choose an action from the list of possible actions that aligns with the goals outlined in the plan, giving precedence to main goal or sub-goals in the order they appear (main goal is highest priority, then sub_goal_1, sub_goal_2, etc.). However, prioritize sub-goals that can be solved by perfroming single action in current situation, like 'take something', over long term sub-goals. 
Actions like "go to 'location'" will move an agent directly to stated location, use them instead of "go_west' type of actions, if the destination you want to move to is further than 1 step away. 
In tasks centered around exploration or locating something, prioritize actions that guide the agent to previously unexplored areas. You can deduce which locations have been visited based on the history of observations and information from your memory module.
Performing same action typically will not provide different results, so if you are stuck, try to perform other actions or prioritize goals to explore the environment.
You may choose actions only from the list of possible actions. You must choose strictly one action.
Write your answer exactly in this json format:

{
  "reason_for_action": "reason"
  "action_to_take": "selected action"
  
}

Do not write anything else.
####
1. Main goal: {main_goal}
2. History of {n_prev} last observations and actions: {observations} 
3. Your current observation: {observation}
4. Information from the memory module that can be relevant to current situation:  {subgraph}
5. Your {topk_episodic} most relevant episodic memories from the past for the current situation: {top_episodic}.
6. Your current plan: {plan0}
7. Yet unexplored exits in the environment: {all_unexpl_exits}

Possible actions in current situation: {valid_actions}
\end{lstlisting}
\end{tcolorbox}

\begin{tcolorbox}[title=Summarization prompt, enhanced jigsaw, breakable, colback=white, colframe=black, sharp corners]
\lstset{
  basicstyle=\tiny\ttfamily,
  breaklines=true,
  columns=fullflexible,
}
\begin{lstlisting}
####
INSTRUCTION:
You are a guide within a team of agents engaging in a text-based game. Your role is to concisely yet thoroughly detail all the essential aspects of the current situation. Ensure that your summary aids in information extraction and facilitates the decision-making process by focusing on pertinent details and excluding extraneous information. Incorporate a strategic outlook in your narrative, emphasizing information integral to forming a tactical plan.

Accurately relay the outcomes of previously attempted actions, as this is pivotal for shaping subsequent choices. Your account will form the sole basis on which the decision-making agents operate; thus, clarity and avoidance of potential confusion are paramount.

Be judicious with your inferences, presenting only well-substantiated information that is likely to be of practical benefit. Your account should be succinct, encapsulated within a maximum of three paragraphs.
####
1. Main goal: {main_goal}
2. History of {n_prev} last observations and actions: {observations} 
3. Your current observation: {observation}
4. Your previous summary: {summary}
Your summary: 
\end{lstlisting}
\end{tcolorbox}

\section{Text-based Games}
\label{app:games}

\begin{figure*}[h]
  \centering
  \includegraphics[width=.8\textwidth]{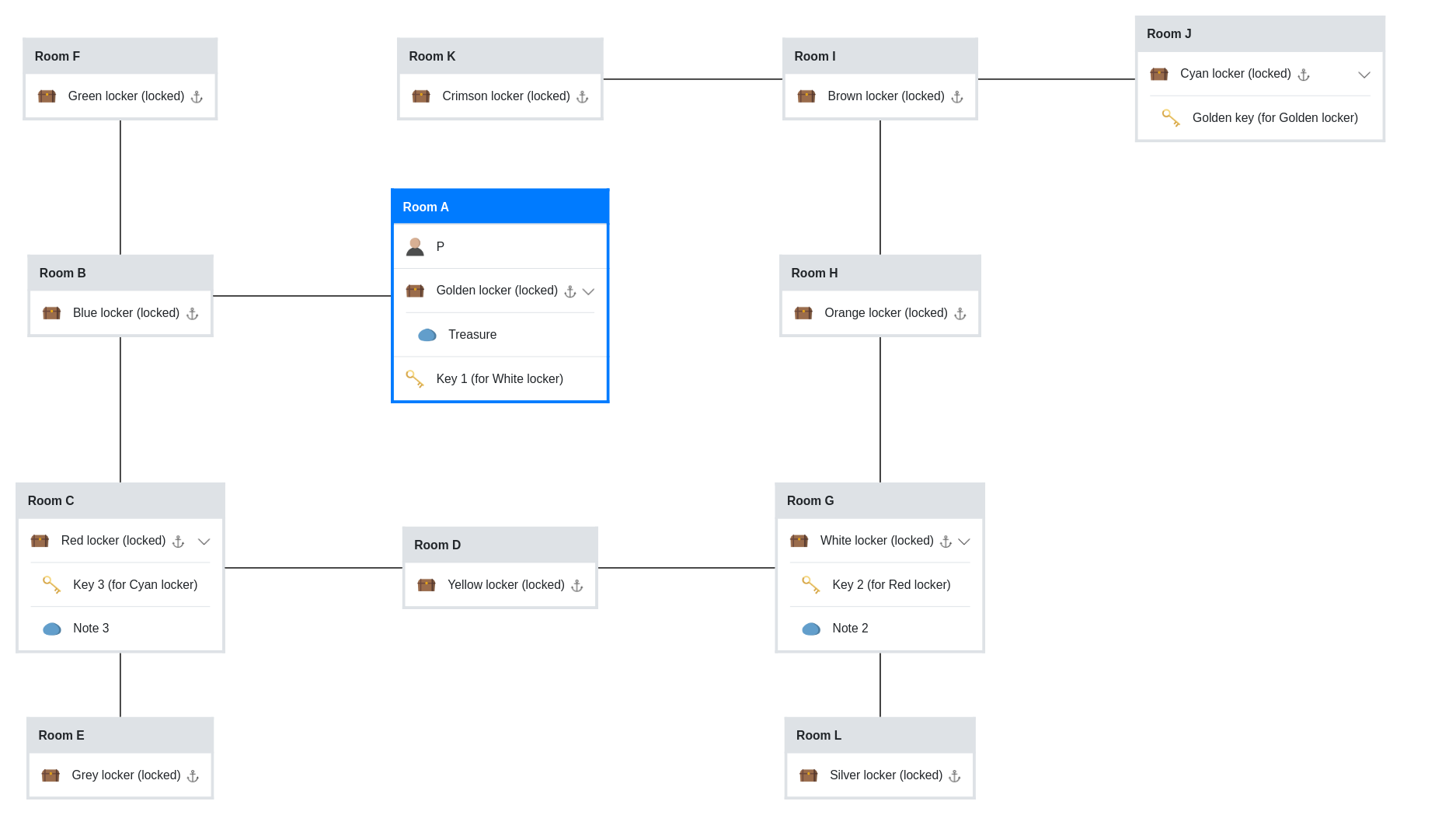}
  \caption{Treasure Hunt environment}
  \label{fig:treasure_hunt}
\end{figure*}

\begin{figure*}[b]
  \centering
  \includegraphics[width=.8\textwidth]{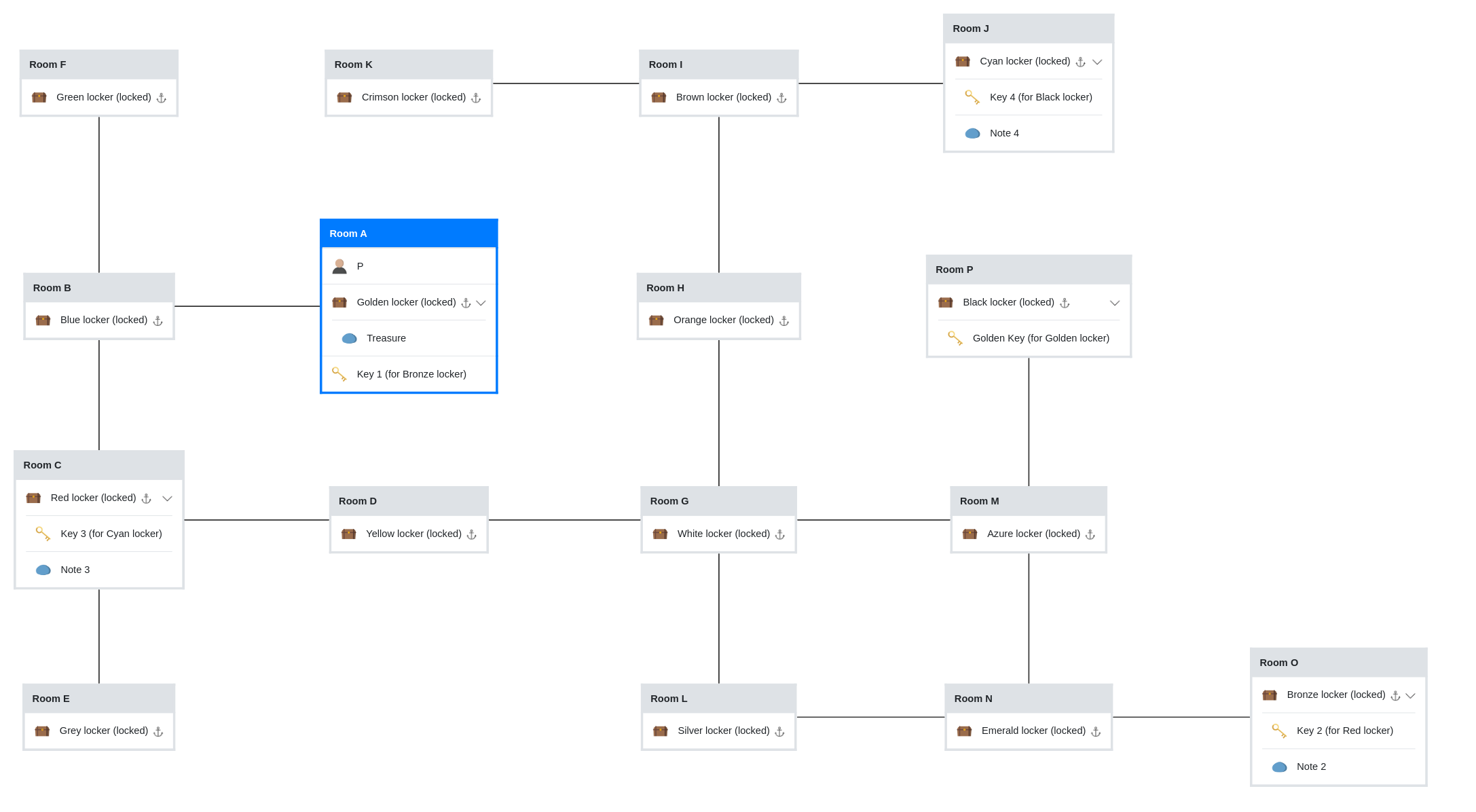}
  \caption{Treasure Hunt hard environment}
  \label{fig:treasure_hunt_hard}
\end{figure*}

\begin{figure*}[h]
  \centering
  \includegraphics[width=.8\textwidth]{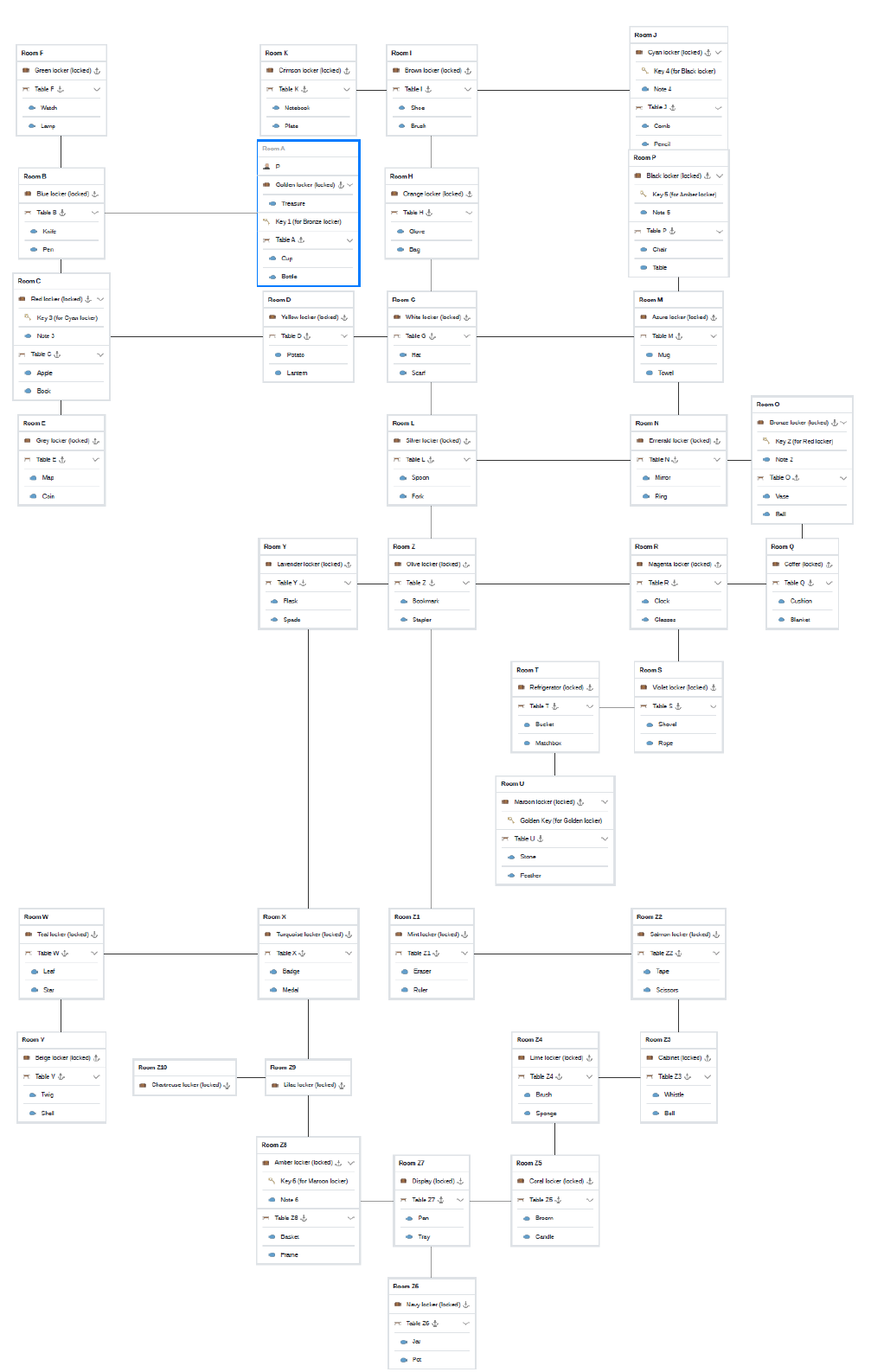}
  \caption{Treasure Hunt hardest environment}
  \label{fig:treasure_hunt_hardest}
\end{figure*}

\textbf{Treasure Hunting.} The primary objective is to unlock the golden locker and retrieve the treasure hidden within. The game consists of rooms each featuring a locker of a different color. Initially, the player finds a key in the first room, along with instructions detailing which specific locker this key can unlock. Each subsequent locker contains another key and a note directing the player to the next key's location, creating a chain of clues and discoveries leading to the golden locker. The easy variation has 12 rooms and 4 keys and hard one has 16 rooms and 5 keys, however second key is significantly harder to find. The hardest variation consist of 36 total rooms, 7 keys and also additional items in each room as noise. 
Agent receives 1 point for picking a key and 1 point for completing the game. Examples of the Treasure Hunt environment are presented in figures \ref{fig:treasure_hunt}, \ref{fig:treasure_hunt_hard}
\ref{fig:treasure_hunt_hardest}

\begin{figure*}[b]
  \centering
  \includegraphics[width=.8\textwidth]{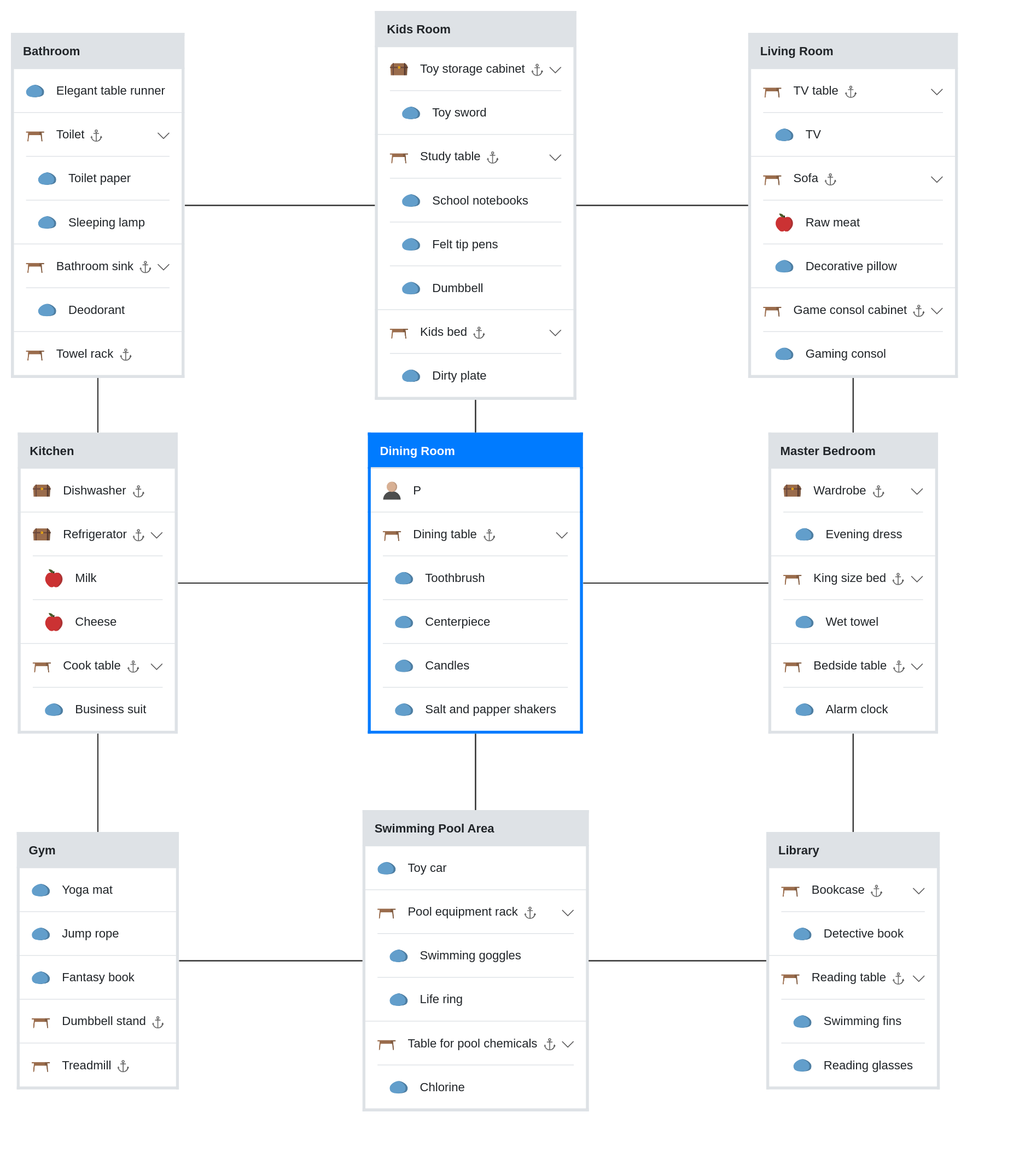}
  \caption{Cleaning environment}
  \label{fig:cleaning}
\end{figure*}

\textbf{Cleaning.} The goal is to clean the house that consists of 9 distinct rooms, each designated for a specific purpose, such as a pool, kitchen, etc. Each room contains items, some of which are misplaced, for example, a toothbrush found in the dining room. There are total of 11 misplaced items. The agent’s objective is to tidy up the house by identifying items that are out of place and returning them to their appropriate locations. This task requires the agent to utilize reasoning skills to determine the proper locations for items and to effectively recall from memory where each item belongs, while simultaneously managing multiple tasks. Agent receives 1 point for picking up every displaced items, 1 point for placing an item in the right location and -1 for manipulation with correctly placed item. Our task is conceptually similar to the TextWorld Commonsense (TWC)  benchmark\cite{murugesan2020textbased}. However, while TWC primarily centers on logical reasoning within one or at most two locations, our emphasis is on environmental exploration and the testing of memory based on past observations. Consequently, we have substantially expanded the number of rooms and items in our setup. Example of the Cleaning environment can be found in figure \ref{fig:cleaning}.

\begin{figure*}[h]
  \centering
  \includegraphics[width=.8\textwidth]{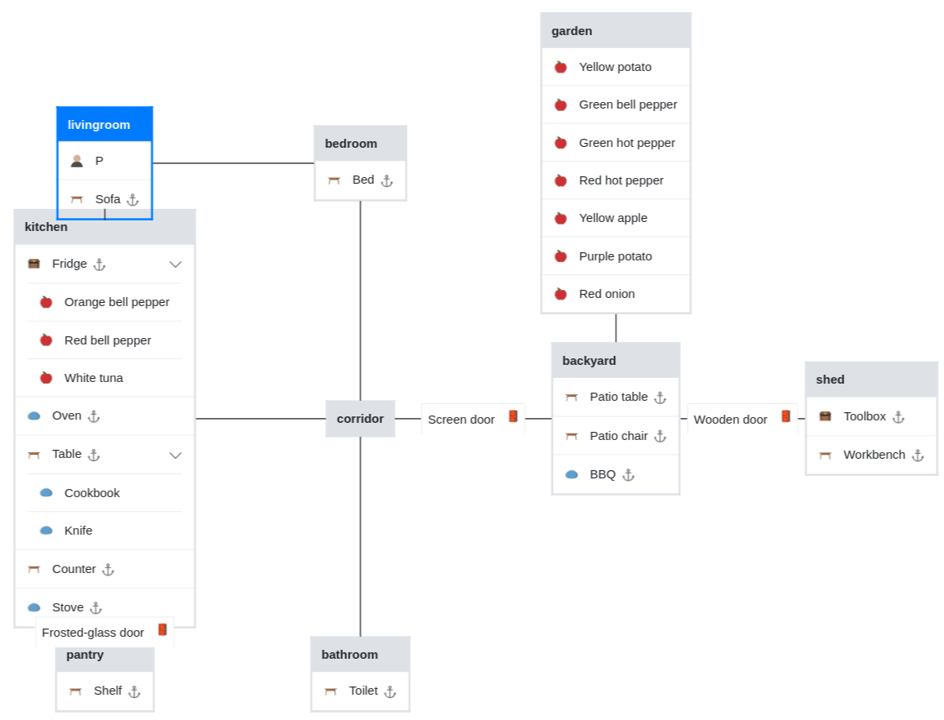}
  \caption{Cooking environment}
  \label{fig:cooking}
\end{figure*}

\begin{figure*}[h]
  \centering
  \includegraphics[width=.8\textwidth]{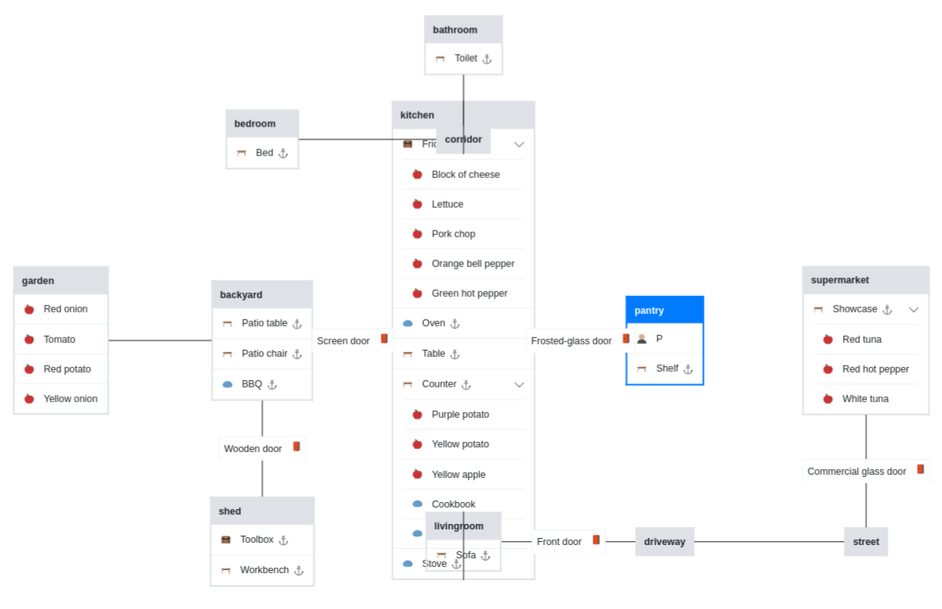}
  \caption{Cooking hard environment}
  \label{fig:cooking_hard}
\end{figure*}

\textbf{Cooking.} The goal is to prepare and consume a meal. First an agent needs to find a recipe that provides detailed instructions, including required ingredients and specific preparation methods such as dicing, slicing, chopping, grilling, frying, and roasting. Medium difficulty task features 9 locations and 3 ingredients and hard task features 12 locations and 4 ingredients.  The agent receives points for correctly selecting ingredients and executing the proper procedures. The game is considered lost if the agent performs any incorrect action or uses an inappropriate tool for a given ingredient. We enhanced the initial observation with instructions and provided explanations for specific actions to tailor the task for LLM models, particularly concerning the appropriate use of household kitchen appliances for different actions with ingredients. For instance, a BBQ should be used for grilling, while a stove is appropriate for frying (see Appendix \ref{app:prompts} for prompt). This allows to test the agent's ability to remember and adhere to instructions. Examples of the Cooking environment are presented in figures \ref{fig:cooking}, \ref{fig:cooking_hard}. 

\textbf{RL comparison benchmark.} We tested our architecture on a variation of the Cooking test from \cite{adhikari2021learning} to compare it with RL baselines. These tasks have 4 levels of difficulty, however, they are significantly simpler than our main tasks, having fewer locations, ingredients, and required actions.
\begin{itemize}
    \item Level 1: 1 ingredient, 1 room, cutting.
    \item Level 2: 1 ingredient, 1 room, cutting + cooking.
    \item Level 3: 1 ingredient, 9 rooms, no processing.
    \item Level 4: 3 ingredients, 6 rooms, cutting + cooking.
\end{itemize}
We tested or agent and raw GPT-4 with full history on 3 randomly generated environments for each task difficulty. 
We slightly adapted the task for the LLM models. We gave the instruction to grill, roast or fry the ingredient using appropriate kitchen tools such as BBQ, oven and stove in the first observation of the game. RL agent can learn this rules by interacting with the environment, however, this is not a commonsense knowledge and without this instruction the game becomes unbeatable for LLM-agents and human players.  While this adaptation should not impact the difficulty level, it's important to note that the environments for comparison between LLM and RL models were not 100\% identical.

\begin{figure*}[h]
  \centering
  \includegraphics[width=.8\textwidth]{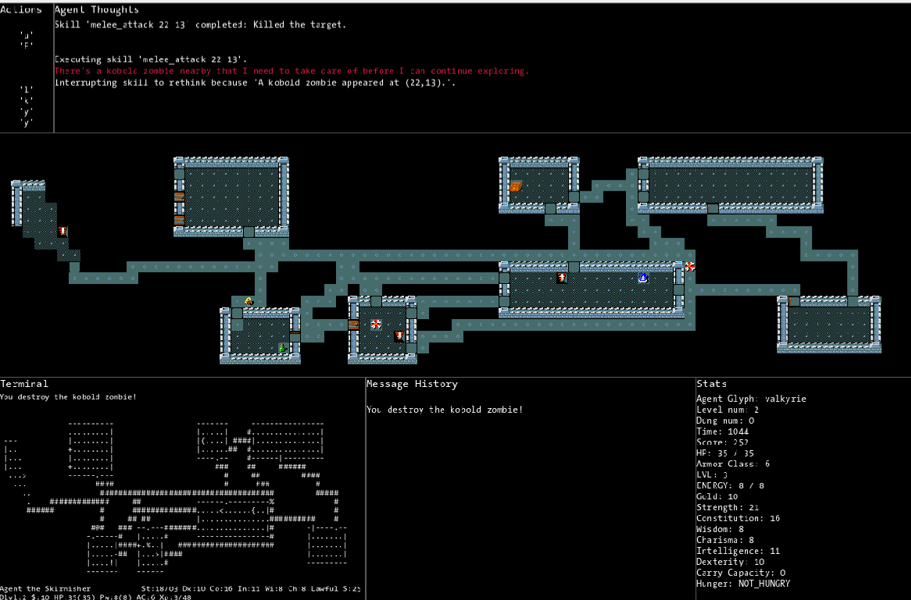}
  \caption{Example of NetHack level}
  \label{fig:nethack_env}
\end{figure*}

\section{Additional graphics and tables}
\label{app:textworld_extra_results}
\begin{figure*}[h!]
  \centering
  \makebox[\textwidth]{\includegraphics[width=1\textwidth]{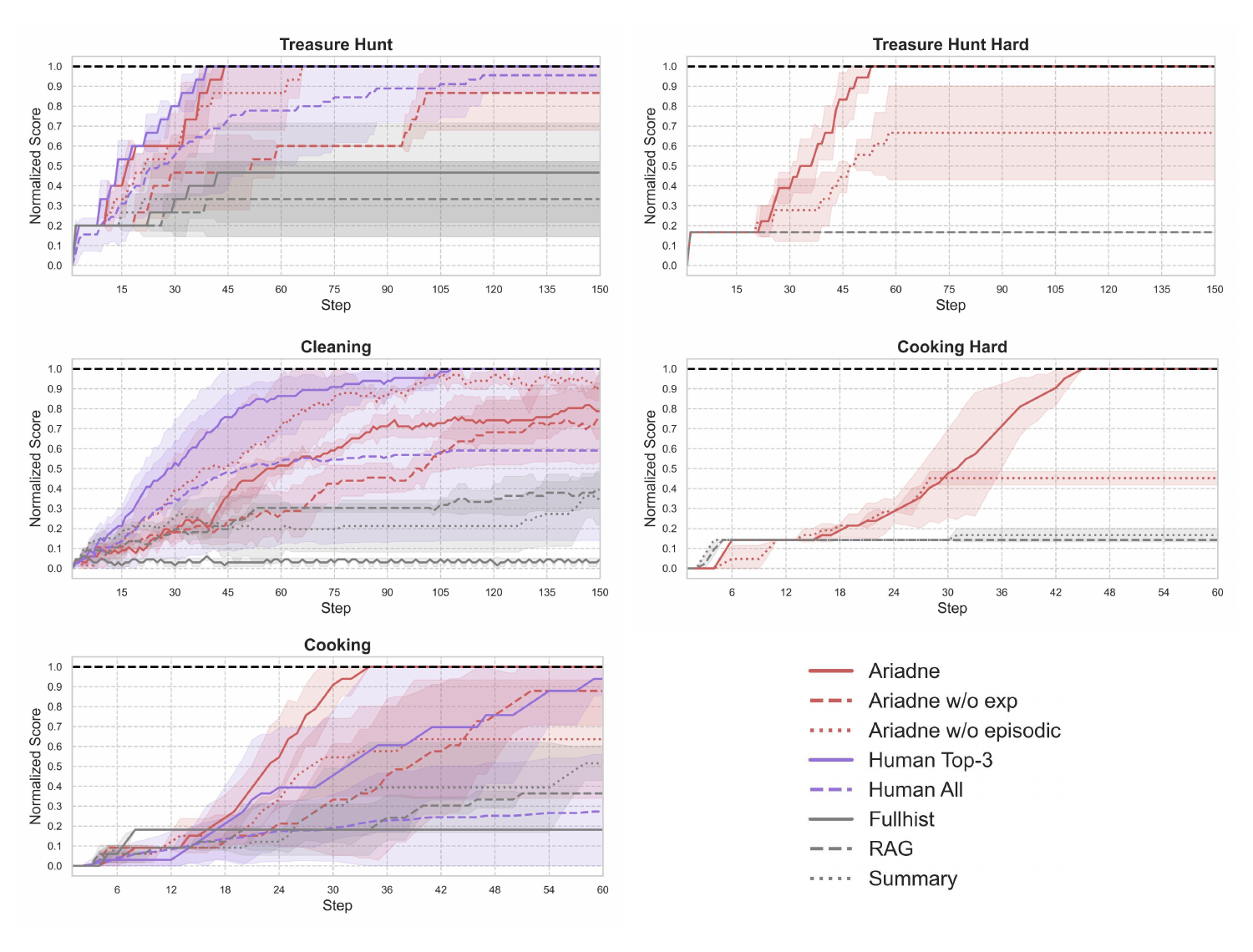}}
  \caption{Performance Dynamics in Test Games. In the Treasure Hunt, the Ariadne agent delivers performance comparable to top players; in the Cleaning task, it falls slightly behind, but in the Cooking, it surpasses top human players in speed. The hard variants of the tasks demonstrate that the quality of Ariadne's performance does not decrease with increasing task difficulty, and also highlight the importance of episodic memory. }
  \label{fig:dynamic}
\end{figure*}

This section presents the step-by-step dynamics of performance of different agents and human players (figure.\ref{fig:dynamic}) . Additionally, this section includes a table summarizing the results of all experiments across three versions of the Treasure Hunt (Medium, Hard, Hardest), three versions of the Cooking (Medium, Hard, Hardest), and the Cleaning (table \ref{env_results_tab}).

\begin{table*}[h!]
    \centering
    \begin{tabular}{l|p{1.5cm}p{1.5cm}p{1.5cm}|p{1.5cm}p{1.5cm}p{1.5cm}|p{1.5cm}}
        \toprule
        & \textbf{Treasure Hunt}  & \textbf{Treasure Hunt Hard} & \textbf{Treasure Hunt Hardest} & \textbf{Cooking} & \textbf{Cooking Hard} & \textbf{Cooking Hardest} & \textbf{Cleaning }\\
        \midrule
        Full History & 0.47 & - & - & 0.18 & - & - & 0.05 \\
        Summary & 0.33 & 0.17 & - & 0.52 & 0.21 & - &0.35\\
        RAG & 0.33 & 0.17 & - & 0.36 & 0.17 & - & 0.39\\
        Reflexion & 0.93 & - & - & \textbf{1.0} & - & - & 0.27\\
        Simulacra & 0.4 & - & - & 0.3 & - & - & 0.7\\
        \midrule
        AriGraph & \textbf{1.0} & \textbf{1.0} & \textbf{1.0} & \textbf{1.0} & \textbf{ 1.0} & 0.65 & 0.79\\
        AriGraph w/o exploration & 0.87 & - & - & 0.87 & - & - & 0.76\\
        AriGraph w/o episodic & \textbf{1.0} & 0.67 & - & 0.64 & 0.45 & - & 0.92\\
        AriGraph LLaMA-3-70B & 0.47 & - & - & 0.67 & - & - & 0.5\\
        \midrule
        Human Top-3 & \textbf{1.0} & - & - & \textbf{1.0} & - & - & \textbf{1.0}\\
        Human All & 0.96 & - & - & 0.32 & - & - & 0.59\\
        \bottomrule
    \end{tabular}
    \caption{\textbf{All normalised scores across all tasks in TextWorld environment.} Based on the results, it is evident that the agent with AriGraph significantly outperforms all baselines and scales well to larger and more complex environments. An important outcome is that our agent demonstrated near-human performance in text-based games, which has not been previously achieved using LLM.}
    \label{env_results_tab}
\end{table*}

\end{document}